\documentclass[journal,comsoc]{IEEEtran}
\usepackage{amsmath,amssymb,amsfonts}
\usepackage{graphicx}
\usepackage{textcomp}
\usepackage{subfig}
\usepackage{tipa}
\usepackage{multirow}
\usepackage{booktabs}
\usepackage{cite}
\usepackage{algorithmic}
\usepackage{xcolor}
\usepackage{moreverb,url}
\usepackage{ dsfont }
\usepackage{comment}
\usepackage[colorlinks,bookmarksopen,bookmarksnumbered,citecolor=red,urlcolor=red]{hyperref}
\usepackage[ruled,vlined,noresetcount]{algorithm2e}
\begin{document}
\title{Loop-box: Multi-Agent Direct SLAM Triggered by Single Loop Closure for Large-Scale Mapping}
\author{M Usman Maqbool Bhutta,~\IEEEmembership{Graduate Student Member,~IEEE}, \\Manohar Kuse, ~\IEEEmembership{Graduate Student Member,~IEEE},
Rui~Fan,~\IEEEmembership{Member,~IEEE}, Yanan Liu,~\IEEEmembership{Graduate Student Member,~IEEE},
Ming~Liu,~\IEEEmembership{Senior Member,~IEEE}\\
\thanks{M. U. M. Bhutta, M. Liu are with the Robotics and Multi-Perception Laboratory in Robotics Institute at the Hong Kong University of Science and Technology, Hong Kong. (email: usmanmaqbool@outlook.com)}
\thanks{R. Fan is with the Jacobs School of Engineering and the Jacobs School of Medicine, the University of California, San Diego, as well as ATG Robotics.}
\thanks{M. Kuse is with Robotics Institute at the Hong Kong University of Science and Technology, Hong Kong.}
\thanks{Y. Liu is with Bristol Robotics Laboratory, University of Bristol, Bristol, BS16 1QY, United Kingdom.}
}
\IEEEtitleabstractindextext{%
\begin{abstract}
 In this paper, we present a multi-agent framework for real-time large-scale 3D reconstruction applications. In SLAM, researchers usually build and update a 3D map after applying non-linear pose graph optimization techniques. Moreover, many multi-agent systems are prevalently using odometry information from additional sensors. These methods generally involve intensive computer vision algorithms and are tightly coupled with various sensors. We develop a generic method for the keychallenging scenarios in multi-agent 3D mapping based on different camera systems. The proposed framework performs actively in terms of localizing each agent after the first loop closure between them. It is shown that the proposed system only uses monocular cameras to yield real-time multi-agent large-scale localization and 3D global mapping. Based on the initial matching, our system can calculate the optimal scale difference between multiple 3D maps and then estimate an accurate relative pose transformation for large-scale global mapping. 
\end{abstract}

}

\maketitle
\IEEEdisplaynontitleabstractindextext
\IEEEpeerreviewmaketitle
\section{Introduction}
\label{sec.introduction}

\begin{figure}[!h]
 \centering
 \includegraphics[width=.95\linewidth]{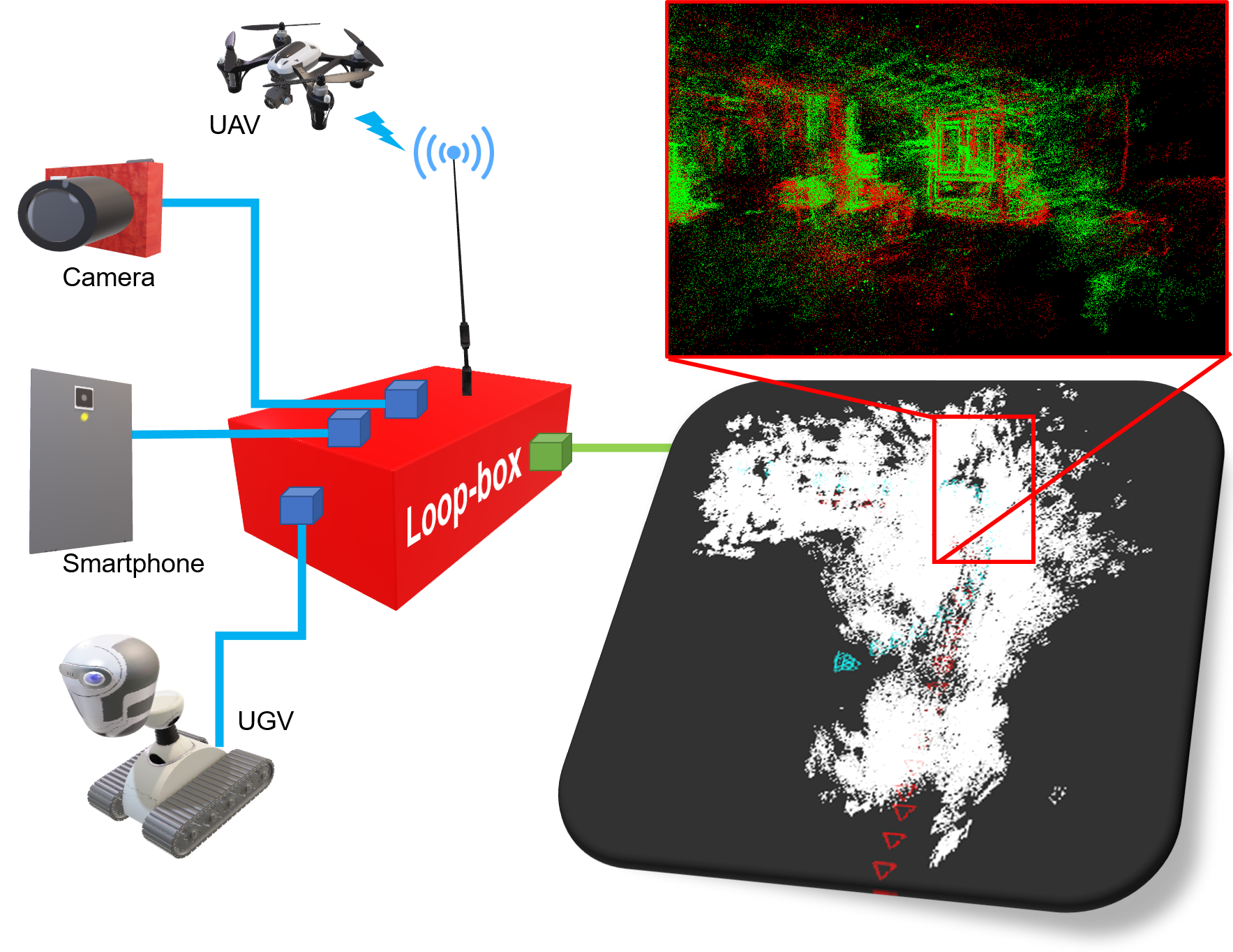}
 \caption{Loop-box assists different cameras integrated into a system for multi-agent SLAM. The poses of the source and target agents are shown in red and cyan, respectively. Their 3-D maps fused using Loop-box are shown in red and green, respectively. }
 \label{fig1}
\end{figure}
In modern times, many researchers have turned their focus towards developing multi-agent simultaneous localization and mapping (SLAM) systems that can merge multiple maps built by each agent connected with the centralized system \cite{schmuck2019ccm,schmuck2017multi}. However, usage of these systems is usually computationally intensive, and it is hard to achieve real-time performance in large-scale applications. Maplab\cite{Schneider2018} has provided a multi-session solution for such SLAM problems. This is useful when the agents lose their relative pose information due to an abrupt change in light, or when someone passes by the camera system. Maplab can easily build a connection between the previous odometry and the new odometry. 

\begin{figure}[!t] 
 \captionsetup[subfigure]{width=.45\linewidth}
 \subfloat[The two agents move in the same direction, and they have several loop closures.]{\label{fig2a}
 \includegraphics[width=.45\linewidth]{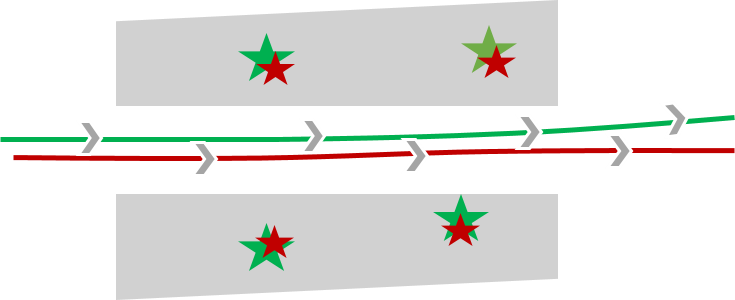}
 }
 \subfloat[The two agents move in the same direction, and they have only one loop closure]{\label{fig2b}
 \includegraphics[width=.45\linewidth]{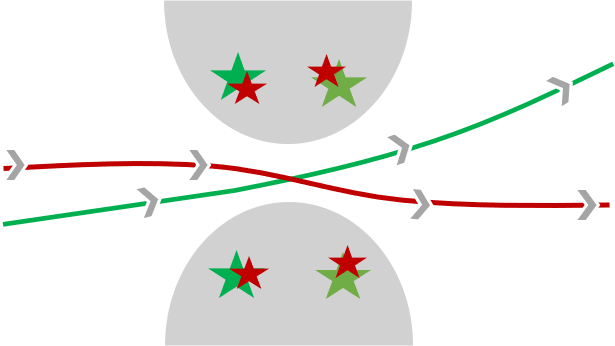}
} \\
 \subfloat[The two agents move in opposite directions, and they have several loop closures.]{\label{fig2c}
 \includegraphics[width=.45\linewidth]{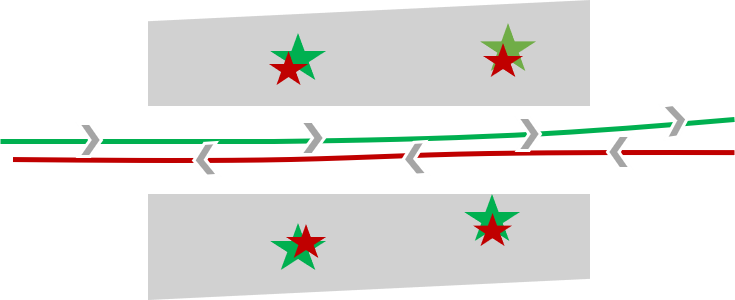}
 }
 \subfloat[The two agents move in opposite directions, and they have one-sided loop closure.]{\label{fig2d}
 \includegraphics[width=.45\linewidth]{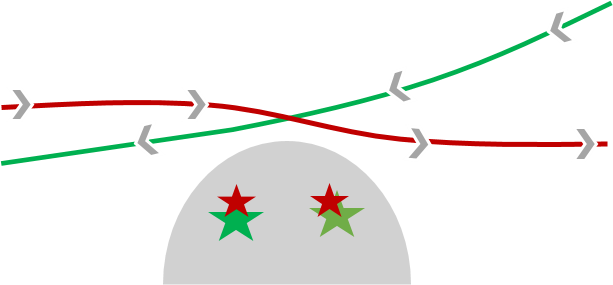}
 }
 \caption{Key challenging states a multi-agent SLAM system may encounter. Red and green lines represent the paths of each of two agents. The landmarks are illustrated with red and green stars. As seen from different agents, a scale difference may occur.}
 \label{fig2} 
\end{figure}

The lack of a state-of-the-art multi-agent SLAM system which can easily connect various robotic platforms also emphasises the importance of developing a ready-to-use framework for multiple agents \cite{Bosse2004}.
Visual-inertial-sensor-based SLAM systems \cite{qin2018vins, Montiel2015, Mur-Artal2017, Mur_Artal_2017} perform excellently for a single agent, but for large-scale 3-D collaborative mapping, a complicated calibration setup along with intensive computing power is essential. In addition to this, the most ambitious problem is to estimate an accurate relative pose transformation between different agents after observing the first loop closure. The system should be sufficiently robust to fuse multiple 3-D maps, regardless of the scale variation between them. An example of a multi-agent system is shown in Fig. \ref{fig1}. It presents the central system that processes the 3-D maps generated by different agents.
Furthermore, the agents can face several key challenging states while moving around. These states are shown in Fig. \ref{fig2}. 

In the present studies, agents are constrained to the loop closures where they are following the same path, as shown in Fig. \ref{fig2a}. Fig. \ref{fig2b} to \ref{fig2d} show real-world scenarios that agents might face to achieve loop closure during a short interval, e.g., when agents are following a same direction and detect a loop closure for a short time. Another intriguing case is where agents are coming from opposite directions, and the system detects a short-duration loop closure. Hence, a robust multi-agent framework is needed to determine the optimum relative pose difference of such agents.
\subsection{Contributions}
\label{sec.contributions}
We develop an intelligent system that can help establish a relative connection between the agents at first interaction. Our proposed framework is capable of working at the first loop closure under the key challenging scenarios mentioned earlier, in Fig. \ref{fig2}. To show the robustness of our system, we use the large-scale direct (LSD) SLAM\cite{Engel2014} system on each agent. Additionally, this SLAM system does not have any scale information or feature point tracking on each end. Fig. \ref{fig1} presents the final 3-D point cloud, which is shown in an enlarged boxed area with a contrasting color scheme. Red shows the transformed source point cloud map, and green is the target point cloud map. The transformed trajectories of both agents inside the fused semi-dense point cloud maps using the proposed method are also shown in Fig. \ref{fig1}. The results demonstrate the efficiency of our presented system after the first loop closure; hence its name is \textit{Loop-box}. The Loop-box system can work collaboratively to make a large-scale 3D-map based on semi-dense or sparse SLAM approaches.
 Our contributions include:
\begin{itemize}
 \item A multi-agent direct SLAM system that can merge 3-D maps with different scales to a global representation;
 \item A robust system that can accurately determine the relative change between multiple robots during a very short interval (Fig. \ref{fig2b}, \ref{fig2d}). In particular, no previous study, to our knowledge, has considered short intervals;
 \item A multi-agent system that collaborates pairwise directly. Therefore, each of the agents can estimate the relative pose transformation and registering of the 3-D maps after the first loop closure;
 \item A signal-loop-closure-based method that welcomes features to direct SLAM methods for large-scale 3-D collaborative mapping. This allows the proposed framework to work effectively in any of the states shown in Fig. \ref{fig2}.
\end{itemize} 
We believe that our multi-agent SLAM is one of the first direct SLAM-based multi-agent schemes that unites a wide variety of use-cases within a single system. We firmly believe that computer vision researchers will use it for intelligent mapping and localization of agents for robotics applications.
\subsection{Paper Organization}
The remainder of this paper is organized as follows. Section \ref{sec.related_work} reviews modern multi-agent systems. In Section \ref{sec.preliminaries}, brief introductory material related to this work is included. In Section \ref{sec.proposed_framework}, we provide the details of the proposed framework. In Section \ref{sec.experimental_results}, the experimental results are illustrated, and the performance of the system is evaluated in discussions in Section \ref{sec.discussions}. Finally, Section \ref{sec.conclusions} concludes the paper and provides recommendations for future work.

\section{Related Work}
\label{sec.related_work}
In the world of robotics applications, the 3-D map has great importance, but a single agent is not sufficient to yield a large 3-D map. Therefore, many robotics scientists have turned their concentrations towards developing multi-agent SLAM systems capable of processing multiple maps. The first such reported work is found in \cite{Bosse2004}, where an efficient large-scale mapping and navigation framework is used to incorporate numerous maps. For large-scale mapping, Deutsch et al.\cite{Deutsch2016} proposed a standard solution of using a multi-robot pose graph SLAM . In 2017, Schneider et al.\cite{Schneider2018} introduced an open-source visual-inertial mapping framework named Maplab . This framework provides a collection of tools, including visual-inertial batch optimization, loop closure detection, and map merging \cite{Schneider2018}. Maplab has proved to be one of the most useful open-source SLAM frameworks in recent years.\\
Schmuck et al.\cite{schmuck2019ccm,schmuck2017multi} presented multi-agent visual-inertial-based SLAM systems that can provide high localization accuracy for a single trajectory, and these approaches can be used in session-based collaborative work. However, due to extensive computation by the centralized system, these systems are limited to a maximum of three agents while mapping the whole environment. Furthermore, Chen et al.\cite{xieyuanli_chen2018} proposed a distributed multi-agent SLAM system that can fulfill a variety of large-scale outdoor tasks without relying on the maps to determine the relative poses.\\
Most of the visual-based \cite{Deutsch2016,gil2010multi} and visual-inertial-based \cite{Schneider2018, Mur_Artal_2017} frameworks can perform well only in the case shown in Fig. \ref{fig2a}, where the two agents move in the same direction and they can observe a several landmarks across the path. However, when the agents do not move in the same direction and their paths intersect once (see Fig. \ref{fig2c} and \ref{fig2d}), the existing multi-agent SLAM systems often fail to merge the 3-D maps. Moreover, the visual-inertial-based frameworks usually require additional sensor calibrations to ensure their performance can give excellent results for multi-agent SLAM system. \\
\section{Preliminaries}
\label{sec.preliminaries}
\subsection{Notations} 
We denote the matrices as a bold face capital letter $\mathbf{R}$ and vectors by a bold face lowercase letter $\bold{t}$. At the $i^{th}$ instant, the robot $ r $-associated keyframe and point clouds are expressed as $\mathbf{K}^i_{r}$ and $ P_{r(i)}^{\mathcal{F}_r^i} $, respectively. 
\begin{figure}[!t]
 \centering
 \includegraphics[width=.95\linewidth]{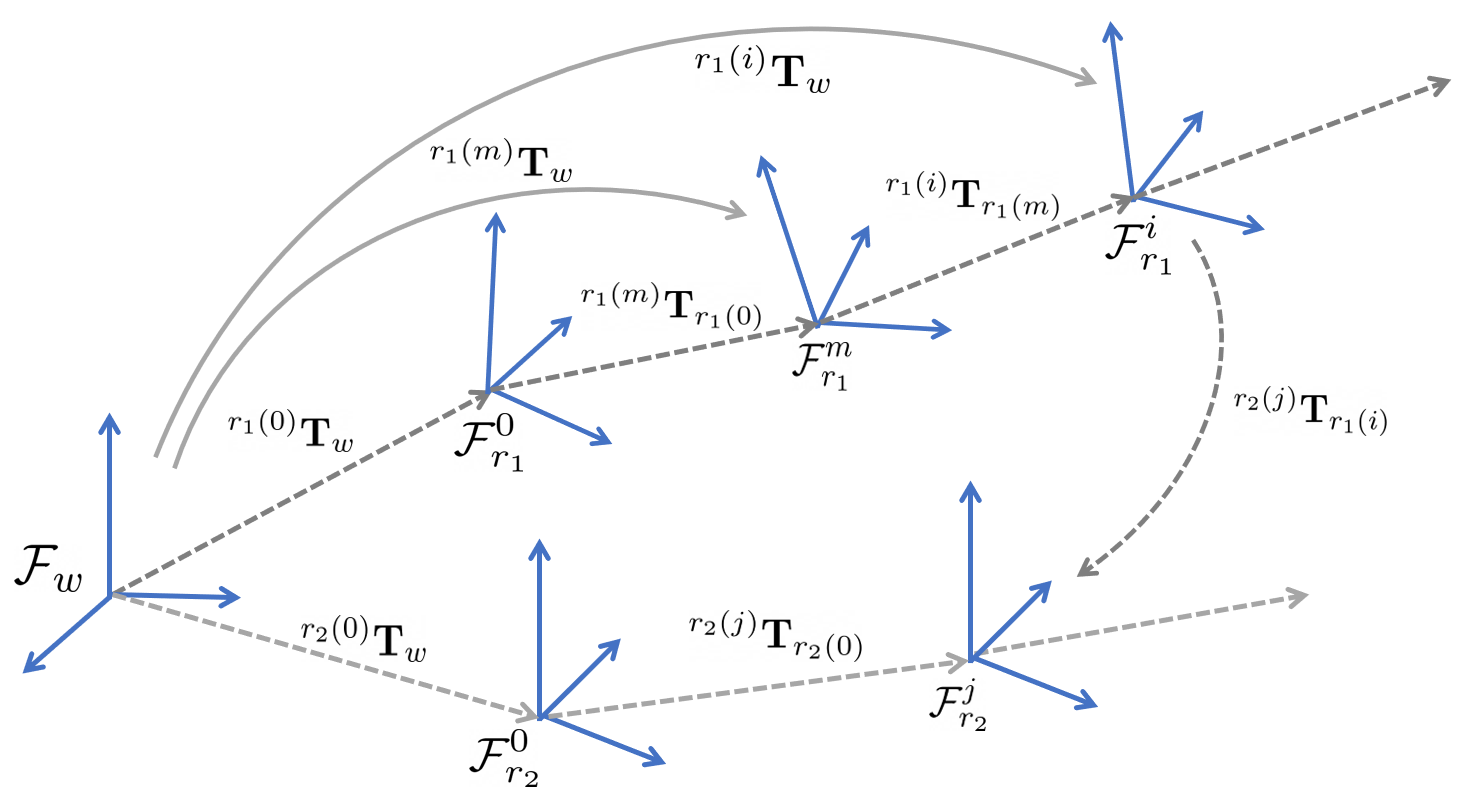}
 \caption{Camera frame transformations relative to the world frame and other agents. The grey continuous-line arrows present the transformation from the world frame. The grey dashed-line arrows represent the relative edges' transformation.}
 \label{fig35}
\end{figure}
\subsection{Transformations} 
We define camera frames by $ \mathcal{F}_{r_\text{ID}} \in \text{SE(3)} \text{ with } \text{ID} \in \mathds{Z}^+$. The operator $^b\mathbf{T}_a(.)$ shows the 3-D rigid body transformation from $ \mathcal{F}_a $ to $ \mathcal{F}_b $ where $^b\mathbf{T}_a \in \text{SE(3)}$. Moreover, $ ^a\mathbf{T}_b $ is the inverse transformation of $ ^b\mathbf{T}_a $. This matrix is further divided into rotation matrix $\mathbf {R}_{ab} \in \text{SO(3)}$ and translation vector $ \mathbf{t}_{ab} \in \mathds{R}^3$. $ ^{r(i)}\mathbf{T}_w $ shows the homogeneous transformation matrix representing the pose of robot $r$ at the $i^{th}$ instant with respect to the world frame $ \mathcal{F}_w $. The transformation notations for a multi-agent system are explained in Fig. \ref{fig35}.
 \begin{figure*}[!t] 
    \captionsetup[subfigure]{width=.65\linewidth}
    \subfloat[Loop-box framework for multi-agent SLAM system. The top part of the diagram shows the Loop-box integration between the source and target agent. All the agents push their keyframes, point clouds, and poses to the system.] {\label{fig3a}
    \includegraphics[width=.70\linewidth]{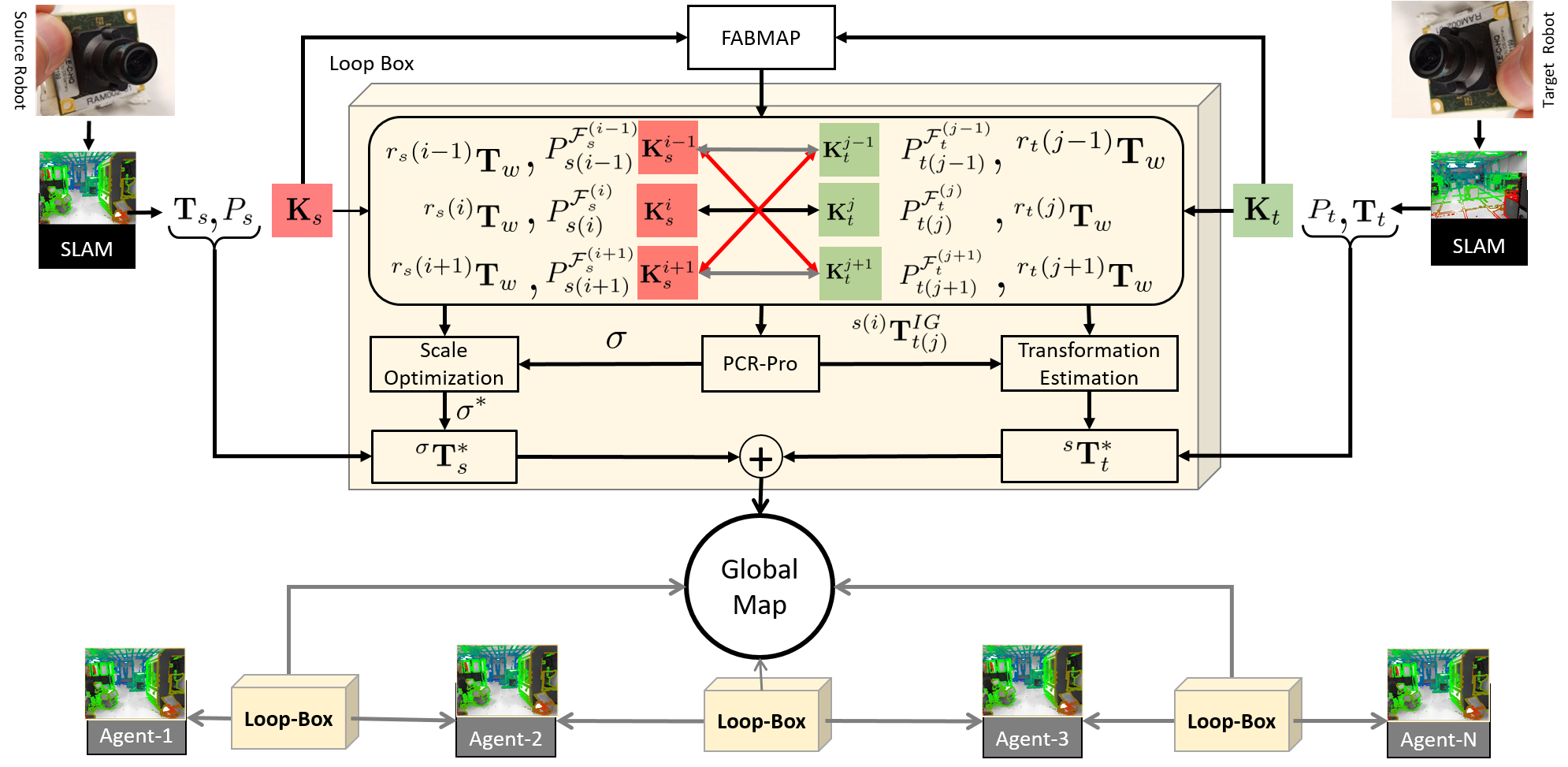}
    }
    \captionsetup[subfigure]{width=.20\linewidth}
    \subfloat[Monocular camera mounted on the top of the UGV.] {\label{fig3b} 
    \includegraphics[width=.26\linewidth]{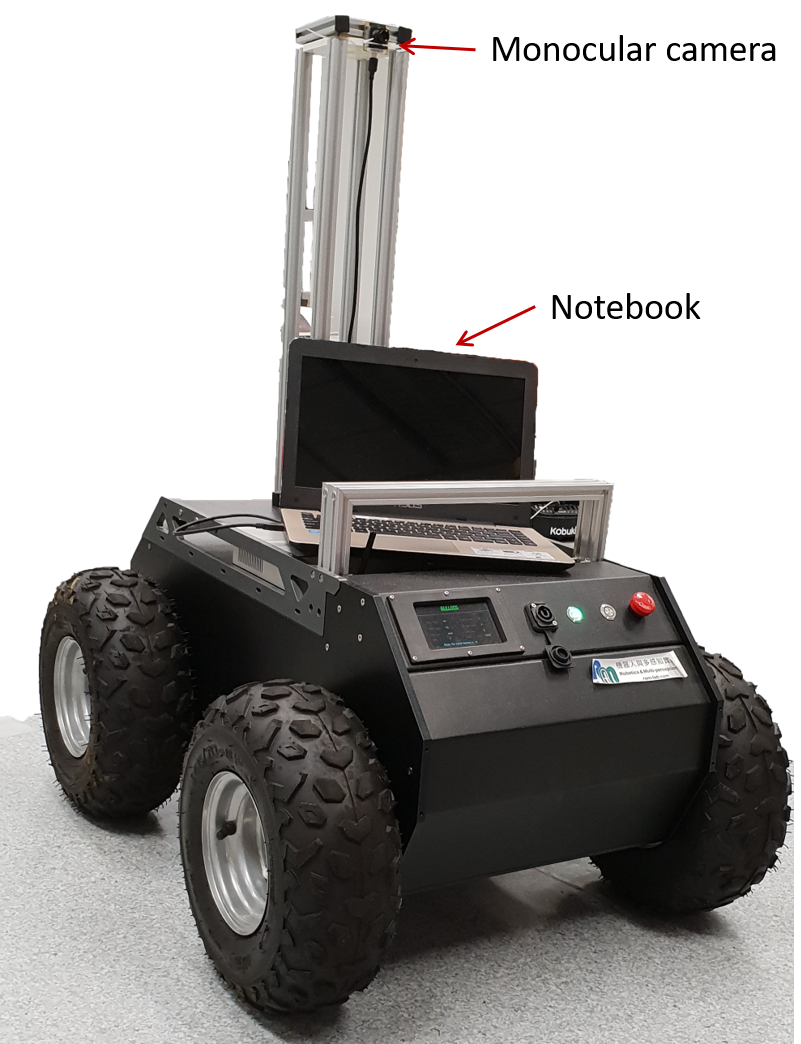}
    }
    \caption{(a) The system overview of the Loop-box framework. Each agent uses a monocular camera and runs LSD-SLAM over it. The keyframes ($\textbf{K}_s$ and $\textbf{K}_t$) are processed for the loop closure detection using FABMAP. The camera frames ($\mathcal{F}_s$, $\mathcal{F}_t$) and point clouds ($P_s$, $P_t$) are transformed using the final scale and relative transformation for the global mapping. (b) The UGV used for the collection of the datasets.}
    \label{fig3} 
   \end{figure*}
Furthermore, $$ \mathcal{F}_b = (^a\mathbf { T }_b)^{-1}. \mathcal{F}_a = \, ^b\mathbf { T }_a. \mathcal{F}_a, $$ where 
$^b\mathbf { T }_a $ transforms the pose of the robot from camera frame $\mathcal{F}_a$ to frame $\mathcal{F}_b$. 
Similarly, transformation $ ^y\mathbf { T }_w = \, ^y\mathbf{T}_x ^x\mathbf { T }_w $ is used to rotate and translate the transformation from $ ^x\mathbf { T }_w $ to $ ^y\mathbf { T }_w $.
A 3-D relative transformation $\textbf{T} \in \text{SE(3)}$ is further divided into scaling ($\sigma$), rotation matrix $\mathbf {R}$ and translation vector $ \mathbf{t}$, which is defined by 
\begin{equation}
\textbf{T} = \begin{pmatrix} \sigma\textbf{R} & \textbf{t} \\ \textbf{0} & 1 \end{pmatrix} \text{ where } \mathbf {R} \in \text{SO(3)}, \mathbf{t} \in \mathds{R}^3 \text{ and } \sigma \in \mathds{R}^+.
\label{eqn-scale}
\end{equation}
\subsection{Information Matrix for Pose Graph SLAM}
For the estimation of uncertainty among edges in pose graph SLAM, the information matrix plays a key role. We calculate the information matrix by taking the inverse transform of a 6x6 covariance matrix in the 6-degrees-of-freedom (DOF) system. We can rewrite the objective function $J$ of the iterative closest point (ICP) as
\begin{equation}
\begin{aligned}
J = {\text{min}}
\sum_{i=1}^{n}{ F }^2,
\end{aligned}
\label{obj_funct_new}
\end{equation}
where $F = \parallel G \parallel$, and $G = \textbf{R}P_i + \textbf{T} - Q_i$. 
For a Jacobian-based objective function, \cite{censi2007accurate} proposed an approximation of the covariance matrix as follows:
\begin{equation}\label{2_eq}
cov(\textbf{x}) \approx \left(\frac{\partial^2 J}{\partial \textbf{x}^2} \right)^{-1} \left( \frac{\partial^2 J}{\partial \textbf{z}\partial \textbf{x}} \right)cov(\textbf{z}) \left( \frac{\partial^2 J}{\partial \textbf{z}\partial \textbf{x}} \right)^T \left(\frac{\partial^2 J}{\partial \textbf{x}^2} \right)^{-1} ,
\end{equation}
where $ \textbf{x} = [x \quad y \quad z \quad a \quad b \quad c]$ and $\textbf{z}$ are the n sets of correspondances $\{P_i, Q_i\}$. In our previous work\cite{Bhutta2018}, we presented an efficient approach, point cloud registration (PCR)-Pro, for aligning and registering multiple 3-D point clouds with different scales. It uses a 3-D-closed-form solution method \cite{Manoj2015} and OpenGV \cite{Kneip2014} for the relative transformation and information matrix computation.

\section{The Loop-box Framework}
\label{sec.proposed_framework}
The Loop-box framework, shown in Fig. \ref{fig3}, can work for both real-time or offline scenarios. It consists of two major components:
\begin{itemize}
 \item For \textbf{distributed multi-agent SLAM}, if agents are capable of exchanging keyframes and detecting the loop closure, Loop-box will help in their relative scale difference computation and estimation of the relative pose transformation.
 \item For \textbf{centralized multi-agent SLAM}, Loop-box processes a few connected keyframes along with their poses and point clouds to merge them.
\end{itemize}
Loop-box is built on the robot operating system (ROS) \cite{quigley2009ros}, and packages are created using the catkin workspace. A ROS-based central system is developed to process the SLAM information of multiple agents. This can be used as an ROS package as well as an extension library to be integrated with any multi-agent SLAM-related applications. The C++11 standard is used along with third-party libraries such as Eigen \cite{eigenweb}, OpenCV \cite{opencv_library}, OpenGV \cite{Kneip2014}, and Ceres \cite{agarwal2012ceres} for programming the whole system. For the detailed visualization and error estimation, we use rviz, ParaView, and EVO \cite{MichaelGrupp}. 
\subsection{Agents Overview}
We name agents as \textbf{slaves} and the server as the \textbf{master}. All the slave robots are equipped with a monocular camera, and a SLAM operation is executed on them individually. In our case, we use the LSD-SLAM \cite{Engel2014} system on each agent. LSD-SLAM assists in providing keyframe, point cloud, and pose information. ROS is used as a core part of the Loop-box method to connect the master with the slaves. 
\subsection{Loop Closure Detection}
The majority of prior research has applied vocabulary building for place recognition\cite{o2011introduction, galvez2012bags}. Sometimes use of CNNs \cite{arandjelovic2016netvlad} has also given promising results. We use FAB-MAP \cite{cummins2011appearance} for place recognition. For instance, our system detects loop closure $LC_{ij}$ among agents at an instant when the source slave is at position $i$, and the target slave is at $j$. Moreover, $ {P}_{s(i)}^{\mathcal{F}_s^i} $ and ${P}_{t(j)}^{\mathcal{F}_t^j} $ will be observed along with the matched keyframes $\mathbf{K}^i_{s}$ and $ \mathbf{K}^j_{t}$, respectively. After detection of the loop closure, the next and previous keyframes are matched to determine the direction of both agents. If both the agents are moving in the same direction, then the previous and next matches are taken, as shown in grey in Fig. \ref{fig3a}. If the opposite direction is detected, then the alternative matches are taken for further computation, as shown in red in Fig. \ref{fig3a}.
\subsection{Scale Information}
After the loop closure detection, the first step is to find out the scale difference. Monocular-camera-based SLAM and direct SLAM approaches lack the provision of scale information. For the best relative transformation between both agents, scale difference $\sigma$ is an essential factor. Usually, it is detected by estimating the volumes of both point clouds. Our previously reported PCR-Pro \cite{Bhutta2018} provides a robust method to deal with such problems by calculating the scale difference and relative transformation between them efficiently.
\subsubsection{Relative transformation of matches at `$LC_{ij}$'}
PCR-Pro estimates the scale difference $\sigma$ and then aligns both point clouds. Moreover, it helps in estimating the good feature points amoung keyframes. The matched SIFT features are utilized to measure the relative transformation between each matching keyframe. In PCR-Pro, we use OpenGV \cite{Kneip2014} with the eight-point algorithm method to find out the relative camera pose $^{s(i)}\mathbf{T}_{t(j)}^{RC}$ and initial guess transformation $ ^{s(i)}\mathbf{T}_{t(j)}^{IG} $. 
\subsubsection{Scale computation using Kalman filter}
After evaluating the relative transformation using the eight-point algorithm, the matched points of the source agent keyframe are transformed to a 3-D world frame of reference of target agent $ \mathcal{F}_t $. We use the Kalman filter to map the 3-D key points to the target agent key points. Based on the alignment, the scale difference $\sigma$ is calculated.
\subsubsection{Optimal scale difference $ \sigma^*$}
In our single loop closure method, more than three consecutive keyframes, including $\mathbf{K}^i_s$ and $\mathbf{K}^j_t$ data, are used, as shown in Fig. \ref{fig3a}. For each match, the scale difference is estimated.

\begin{algorithm}[!t]
 \SetKwInput{KwData}{Input}
 \SetKwInput{KwResult}{Output}
 \KwData{Matched keyframes $\mathbf{K}_{r_{ID}} = \{\mathbf{K}^i_{s},\mathbf{K}^j_{t}\}$ , poses $^w\mathbf{T}_{r_{ID}} = \{^w\mathbf{T}_{s}, \, ^w\mathbf{T}_{t} \}$, point clouds $P_{r_{ID}}^\mathcal{F} = \{P^{\mathcal{F}_s^i}_{s(i)},P^{\mathcal{F}_t^j}_{t(j)}$\} with $i,j \in \mathds{Z}^+$ }
 \KwResult{Optimal scale difference $\sigma^*$, initial guess relative transformation $^{si}\mathbf{T}_{ti}^{IG}$ }
 initialization\;
 \For{$z = -1:1$} {
 $ P_{s(i+z)}^{\mathcal{F}_w} = \,^w\mathbf{T}_{s(i+z)} (P^{\mathcal{F}_{s(i+z)}}_{s(i+z)}) $ \;
 $ P_{t(j+z)}^{\mathcal{F}_w} = \,^w\mathbf{T}_{t(j+z)} (P^{\mathcal{F}_{t(j+z)}}_{t(j+z)}) $ \;
 \SetKwFunction{FMain}{ PCR-Pro \cite{Bhutta2018}}
 \SetKwProg{Fn}{Function}{:}{}
 \Fn{\FMain{$\mathbf{K}_{r_{ID}}$,$P_{r_{ID}}^{\mathcal{F}_w}$}}{
 Estimate volume ratio $r_{vol}$ of $P_{s(i+z)}^{\mathcal{F}_w}, P_{t(j+z)}^{\mathcal{F}_w}$ \;
 $^{s(i+z)}\mathbf{T}_{t(j+z)}^{RC} \longleftarrow \gamma_z \longleftarrow \mathbf{K}^{i+z}_{s},\mathbf{K}^{j+z}_{t} $ \; 
 $ \sigma_z \longleftarrow ^{s(i+z)}\mathbf{T}_{t(j+z)}^{RC} , \gamma_z, ^w\mathbf{T}_{s(i+z)} , ^w\mathbf{T}_{t(j+z)} $\;
 $^{s(i+z)}\mathbf{T}_{t(j+z)}^{IG} \longleftarrow \sigma_z, P_{s(i+z)}^{\mathcal{F}_w},P_{t(j+z)}^{\mathcal{F}_w} $ \;
 \KwRet $\sigma_z, ^{s(i+z)}\mathbf{T}_{t(j+z)}^{IG} $ \; }
 }
 \eIf{$r_{vol} > 0.5$}{
 $\Delta^* = 5 $\;
 \For{$x = -1: 1$}{
 \For{$y = -1: 1$}{
 \If{$x \neq y$}{
 $ \Delta = |\sigma_x - \sigma_y| $\;
 \If{$\gamma^* < \gamma_{xy}$ \&\& $\Delta^* > \Delta$ \&\& $\Delta^* \neq 0$ }{
 $\sigma^* = avg(\sigma_x,\sigma_y)$ \;
 $\Delta^* = \Delta$ \;
 $\gamma^* = \gamma_{xy}$ \;
 }
 }
 }
 }
 }{ $\sigma^* =\sigma_{xy=00} $ \; } 
 \caption{Finest Tuning for Optimal Scale Estimation}
 \label{algo1}
\end{algorithm}
Let us assume $\gamma_{ij}$ is the number of matched keypoints among two keyframes, $i$ and $j$. These matches yield a distinct scale difference $\sigma_{ij} $ depending on the number of matched keypoints $\gamma_{ij}$. The optimal scale difference $\sigma^*$ will be
\begin{equation}
\begin{aligned}
\sigma^* &= \operatorname*{argmax}_{\gamma} \frac{1}{2}|\gamma(\sigma_{ij}), \gamma(\sigma_{i'j'}) | ,
\end{aligned}
\label{eqn1}
\end{equation} 
where $\sigma_{ij} $ and $\sigma_{i'j'}$ are the two nearest points such that\\
\begin{equation}
\begin{split}
\quad |\sigma_{ij} - \sigma_{i'j'}| \leq \Delta^* \quad \forall \quad i, j, i' \text{ and } j' \in \mathds{Z}^+, \quad \Delta^* \in \mathbb{R}.
\end{split}
\label{eqn2}
\end{equation} 
The estimation is further explained in Algorithm \ref{algo1}.
\subsection{Relative Pose Transformation}
Point cloud information assists further in finding the relative pose transformation of slave robots. After getting the optimal scale difference $\sigma^*$, the scale factor is multiplied by the point cloud to make both point clouds of equal scale. By multiplying $\sigma^*$ with the rotation matrix $\mathbf{R}$ in Equation \ref{eqn-scale}, we obtain the scaling transformation $^{\sigma}\mathbf{T}^*_{s}$.
For instance, if $P^{\mathcal{F}_s^i}_{s(i)}$ is smaller than $P^{\mathcal{F}_t^j}_{t(j)}$, then $P^{\mathcal{F}_s^i}_{s(i)}$ will be scaled by $\sigma^*$ as follows:
$$^*P^{\mathcal{F}_s^i}_{s(i)} = \, ^{\sigma}\mathbf{T}^*_{s}(P^{\mathcal{F}_s^i}_{s(i)}).$$
\subsubsection{Alignment of point clouds}
After scaling the point cloud of source agent $P^{\mathcal{F}_s^i}_{s(i)}$ by $\sigma^*$, both the point clouds $P^{\mathcal{F}_t^j}_{t(j)}$ and $P^{\mathcal{F}_s^i}_{s(i)}$ are aligned before applying the ICP. Therefore, we transform the point cloud $P^{\mathcal{F}_t^j}_{t(j)}$ from $\mathcal{F}^j_t$ to $\mathcal{F}_w$:
\begin{equation}
\begin{aligned}
P^{\mathcal{F}_w}_{t(j)}= \, ^{t(j)}\mathbf{T}_w^{-1}(P^{\mathcal{F}_t^j}_{t(j)}) = \, ^w\mathbf{T}_{t(j)}(P^{\mathcal{F}_t^j}_{t(j)}).
\end{aligned}
\label{eqn5}
\end{equation}
Then we transform the point cloud $P^{\mathcal{F}_t^j}_{t(j)} $ from $\mathcal{F}_{w}$ to $\mathcal{F}_{s}^i$:
\begin{equation}
\begin{aligned}
P^{\mathcal{F}_{s}^i}_{t(j)} = \, ^{s(i)}\mathbf{T}_{w}(P^{\mathcal{F}_w}_{t(j)}).
\end{aligned}
\label{eqn6}
\end{equation}
Finally, initial guess transformation is applied to align them further with $ ^*P^{\mathcal{F}_s^i}_{s(i)}$:
\begin{equation}
\begin{aligned}
^+P^{\mathcal{F}_s^i}_{t(j)} = \,^{s(i)}\mathbf{T}_{t(j)}^{IG}(P^{\mathcal{F}_s^i}_{t(j)}).
\end{aligned}
\label{eqn7}
\end{equation}
\subsubsection{Transformation Computation}
After transforming $P^{\mathcal{F}_t^j}_{t(j)}$ to $ ^+P^{\mathcal{F}_s^i}_{t(j)} $, both $^+P^{\mathcal{F}_s^i}_{t(j)} $ and $ ^*P^{\mathcal{F}_s^i}_{s(i)} $ are aligned and are almost the same scale. 
By using the filtered point clouds of both agents, the libpointmatcher \cite{Pomerleau2011} approach is applied to find the final transformation $ ^{s(i)}\mathbf{T}_{t(j)}^{\mathcal{F}_{trans}}$, where the whole fusion converges to the global minima. We use the point-to-point matrix-based error method. The system applies this transformation to the target point cloud that fuses with the source point clouds and gives excellent map merging.
\subsection{Update of Poses and 3-D Map}
If we write the compact form of all the transformations, the final optimal transformation can be written as
\begin{equation}
\begin{aligned}
^{s}\mathbf{T}^*_{t} = \,^{s(i)}\mathbf{T}_{t(j)}^{\mathcal{F}_{trans}} * ^{s(i)}\mathbf{T}_{t(j)}^{IG} * ^{s(i)}\mathbf{T}_{w} * ^{t(j)}\mathbf{T}_w^{-1}.
\end{aligned}
\label{eqn10}
\end{equation}
For the global mapping, all the poses and 3-D maps are updated using the $^{\sigma}\mathbf{T}^*_{s}$ and $^{s}\mathbf{T}^*_{t}$ transformation to the source and target slaves, respectively, and the multi-agent SLAM starts working as shown in Fig. \ref{fig_ri_results}.

\section{Loop-box Results and Comparison}
\label{sec.experimental_results}
This work addresses the compatibility and flexibility of multi-agent systems by introducing a robust and highly relaxed environment for a multi-agent SLAM framework. In comparison to the existing systems, Loop-box can localize the agents and fuse their maps effectively without using the feature keypoints' history. It efficiently performs map merging and reliable re-localization of agents at the first loop closure.
\subsection{Datasets}
In this paper, we assess the performance of Loop-box on five different datasets. The \textit{Ri} dataset consists of the indoor movements of two agents but the indoor environment is feature-rich as compared to the outdoor environment. Therefore, we further test our framework on four outside datasets \textit{Academic building} and \textit{Parking 1, 2 and 3}. The \textit{Ri} and \textit{Parking 1} datasets were taken using a monocular camera connected with a laptop, carried by a human as an agent. Meanwhile, the \textit{Academic building}, \textit{Parking 2}, and \textit{Parking 3} bag files were captured using an unmanned ground vehicle (UGV), which is shown in Fig. \ref{fig3b}. For the multi-agent scenario, the target agent of the \textit{Parking 2} dataset is the source agent of the \textit{Parking 3} dataset. 
\subsection{Quantitative and Qualitative Analysis}
We qualitatively and quantitatively evaluate the Loop-box framework for both online and offline testing. We record the bags files at different times to capture various conditions. For instance, some objects such as cars are present in the \textit{Parking 2} dataset, whereas they are missing in the \textit{Parking 3} dataset. One of the numerous benefits of using the one loop-closure-based method is that it can easily estimate relative transformation in those environments where agents meet for a short interval.
\begin{figure}[!t] 
 \captionsetup[subfigure]{width=.48\linewidth}
 \subfloat[Original visual odometry of both agents start from the origin.]{\label{fig_ri_resultsa}
 \includegraphics[width=.48\linewidth]{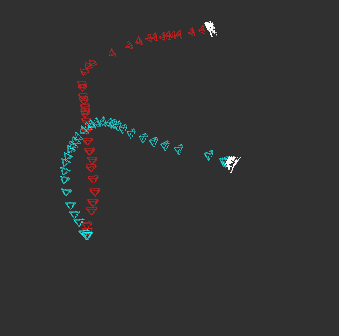}
 }
 \subfloat[Visual odometry poses of the target robot are synchronized to the source robot's frame of reference.]{\label{fig_ri_resultsb}
 \includegraphics[width=.48\linewidth]{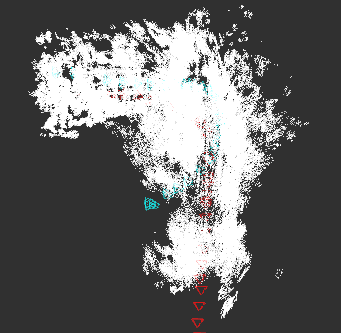}
 } \\
 \captionsetup[subfigure]{width=.95\linewidth}
 \subfloat[Detailed insight of excellent 3-D map merging using Loop-box.]{\label{fig_ri_resultsc}
 \includegraphics[width=.98\linewidth]{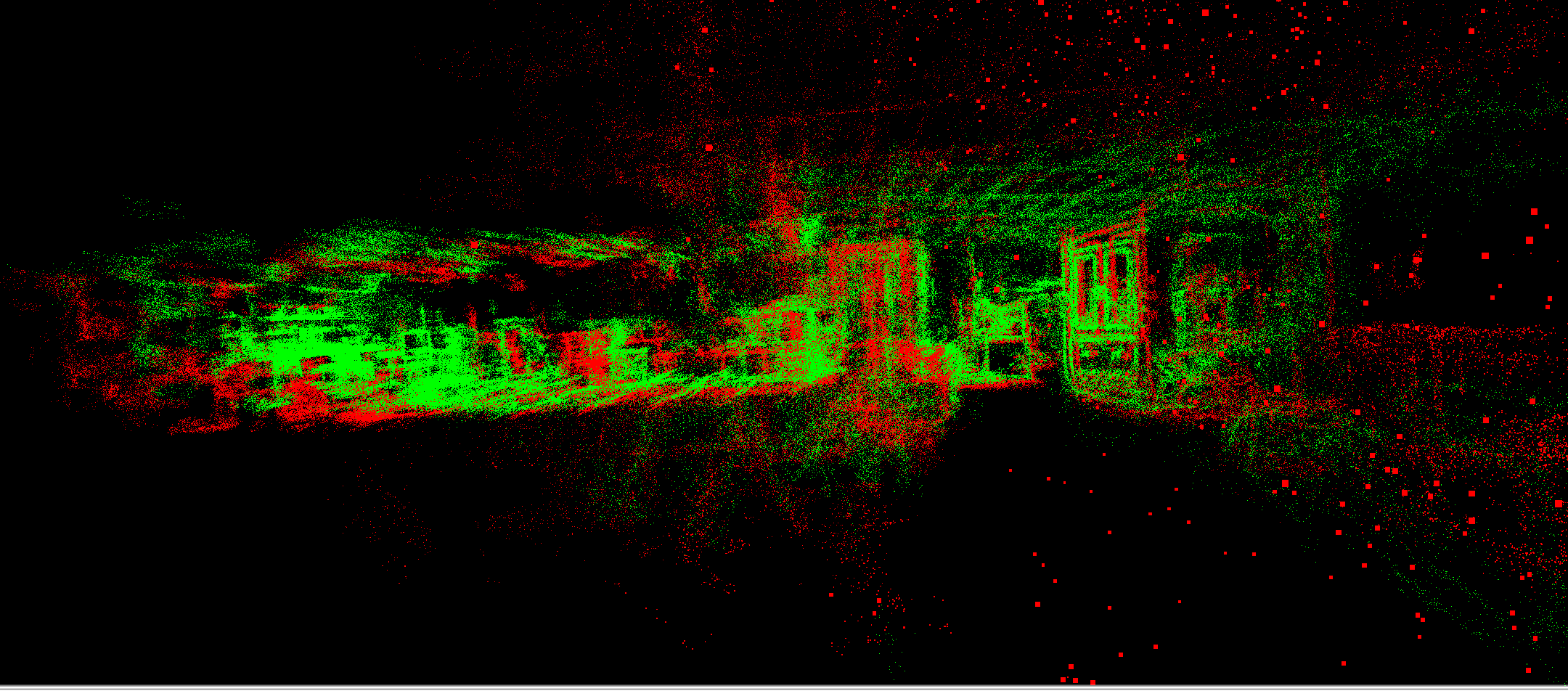}
 }
 \caption{\textit{Ri} dataset results: Live multi-agent SLAM after applying the Loop-box method. The poses and 3-D maps of the source slave are shown in red, while the target slave poses are shown in cyan, with the 3-D map in green.}
 \label{fig_ri_results} 
\end{figure}
\begin{figure}[!t] 
 \centering
 \captionsetup[subfigure]{width=.70\linewidth}
 \subfloat[Visual odometry poses of the transformed target agent to the source agent's frame of reference.]{\label{fig7_ri_posea}
 \includegraphics[width=.75\linewidth]{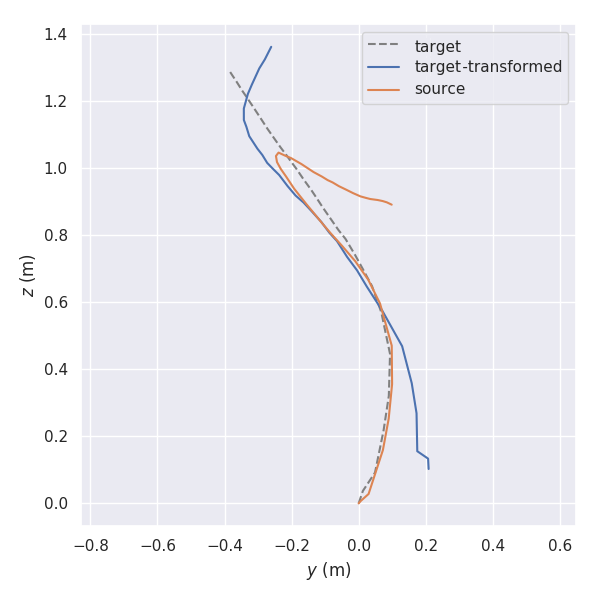}
 } \hfil
 \captionsetup[subfigure]{width=.45\linewidth}
 \subfloat[After the transformation, visual odometry comparison on each axis.]{\label{fig7_ri_poseb}
 \includegraphics[width=.48\linewidth]{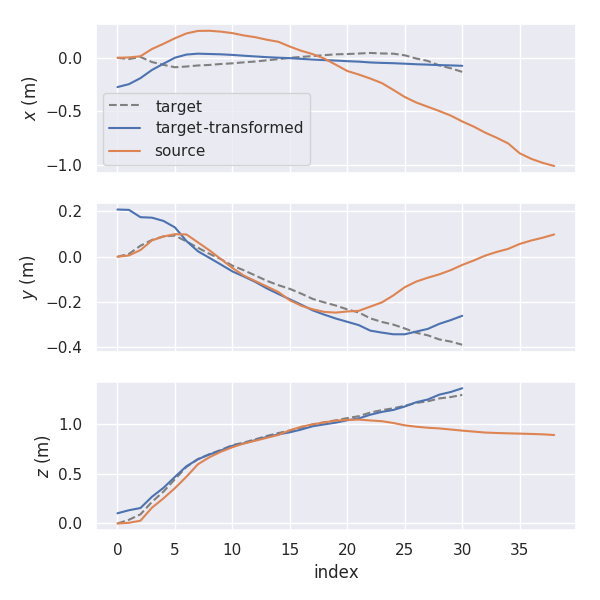}
 }
 \subfloat[Roll, pitch, and yaw results after the transformation.]{\label{fig7_ri_posec}
 \includegraphics[width=.48\linewidth]{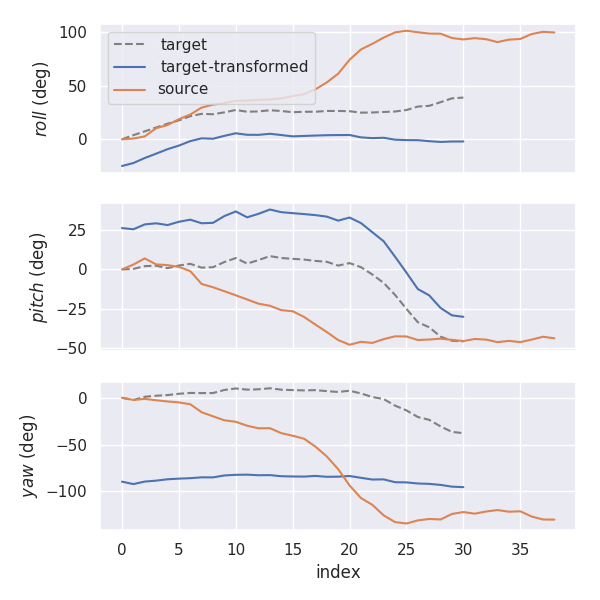}
 }
 \caption{Relative trajectory transformation (RTT) for the enhanced loop closure evaluation on the \textit{RI} dataset. The final trajectories of the source and target agents are shown in orange and blue, respectively.}
 \label{fig7_ri_pose} 
\end{figure}
\subsubsection{Qualitative Results}
The \textit{Ri} dataset was recorded inside the Robotics Institute of HKUST. We captured it using a monocular camera connected to a laptop at a normal speed of human motion. We apply the $^{\sigma}\mathbf{T}^*_{s}$ and $^{s}\mathbf{T}^*_{t}$ transformation estimated in the last section to the poses and point clouds of both agents. It transforms the poses of the target slave to the source slave frame of reference $\mathcal{F}_s$, as shown in Fig. \ref{fig_ri_resultsb}. The SLAM and map merging results are shown in Fig. \ref{fig_ri_results}. 
The original trajectory of the target agent starts with the source agent from the origin $[0,0,0]$. For the multi-agent case, SLAM systems have an optimal relative transformation to another agent's frame of reference.

After applying the transformation estimated using Loop-box, the relative trajectory transformation (RTT) is as shown in Fig. \ref{fig7_ri_posea}.
We further analyze the behavior from each axis, as shown in Fig. \ref{fig7_ri_poseb}. In each direction, we can observe that they both start at the same point. Furthermore, there is a significant movement of the target agent in the x- and z-direction. But in the y-direction, few variations of poses are detected, as the camera height is almost the same throughout the agents' movement. We also observe the changes in terms of yaw, pitch, and roll, as shown in Fig. \ref{fig7_ri_posec}. After applying the Loop-box method, the target rotation is corrected and transformed according to the source agent reference frame. Large variations of poses are observed in the roll and pitch axis. Furthermore, about a two-times shift in the yaw axis is also detected. 
\begin{table*}[]
 \centering
 \renewcommand{\arraystretch}{1.5}
 \caption{Relative pose RMSE of the proposed method compared to PCR-Pro and pose graph optimization.}
 \label{table_1}
 \begin{tabular}{||c|c|c|c|cc|cc|cc|cc|cc||}
 \hline
 \multirow{2}{*}{\textbf{Dataset}} & \multirow{2}{*}{\textbf{Type}} & \multirow{2}{*}{\textbf{Agents}} & \multirow{2}{*}{\textbf{Direction}} & \multicolumn{2}{c|}{\begin{tabular}[c]{@{}c@{}}\textbf{Total Time} \\ \textbf{Duration (sec)}\end{tabular}} & \multicolumn{2}{c|}{\textbf{Scale}} & \multicolumn{2}{c|}{\textbf{PCR-Pro} \cite{Bhutta2018}} & \multicolumn{2}{c|}{\begin{tabular}[c]{@{}c@{}}\textbf{Pose Graph} \\ \textbf{Optimization }\cite{kummerle2011g}\end{tabular}} & \multicolumn{2}{c||}{\textbf{\begin{tabular}[c]{@{}c@{}}Loop-box\\ (Proposed)\end{tabular}}} \\ \cline{5-14} 
 & & & & \begin{tabular}[c]{@{}c@{}}Source \\ Agent\end{tabular} & \begin{tabular}[c]{@{}c@{}}Target \\ Agent\end{tabular} & \begin{tabular}[c]{@{}c@{}}Estimation \\ Time (sec)\end{tabular} & \begin{tabular}[c]{@{}c@{}}Scale \\ Difference\end{tabular} & \begin{tabular}[c]{@{}c@{}}Time \\ (sec)\end{tabular} &  \begin{tabular}[c]{@{}c@{}}RMSE \\ (m)\end{tabular} & \begin{tabular}[c]{@{}c@{}}Time \\ (sec)\end{tabular} &  \begin{tabular}[c]{@{}c@{}}RMSE \\ (m)\end{tabular} & \begin{tabular}[c]{@{}c@{}}Time \\ (sec)\end{tabular} & \begin{tabular}[c]{@{}c@{}}RMSE \\ (m)\end{tabular} \\ \hline \hline
 Ri & Indoor & 2 & Same & 53.7 & 69 & 0.98 & 2.4922 & 7.6 & 0.1503 & 5.1 & 0.0903 & \textbf{2.53} & \textbf{0.0290} \\ \hline
 \begin{tabular}[c]{@{}c@{}}Academic\\ building\end{tabular} & Outdoor & 2 & Same & 73 & 74 & 0.89 & 2.83519 & 8.8 & 1.2453 & 5.5 & 0.9832 & \textbf{2.93} & \textbf{0.0323} \\ \hline
 \begin{tabular}[c]{@{}c@{}}Parking 1 \end{tabular} & Outdoor & 2 & Same & 104 & 91 & 0.94 & 1.7857 & 7.4 & 1.5594 & 5.04 & 0.0883 & \textbf{2.47} & \textbf{0.0486} \\ \hline
 \begin{tabular}[c]{@{}c@{}}Parking 2 \end{tabular} & \multirow{2}{*}{Outdoor} & \multirow{2}{*}{3} & Opposite & 122 & 117 & 0.92 & 1.2786 & 9 & 1.6 & 5.57 & 0.207 &\textbf{ 3} & \textbf{0.175} \\ \cline{1-1}\cline{4-14} 
 \begin{tabular}[c]{@{}c@{}}Parking 3 \end{tabular}& & & Same & 122 & 112 & 0.867 & 1.15 & 11 & 0.3613 & 6.24 & 0.3073 & \textbf{3.67} & \textbf{0.0712} \\ \hline
 \end{tabular}%
\end{table*}
\subsubsection{Quantitative Results}
\begin{figure*}[!t] 
 \captionsetup[subfigure]{width=.40\linewidth}
 \subfloat[Relative pose error after applying PCR-Pro to the parking datasets.] {\label{fig-ape-pcr-pro} 
 \includegraphics[width=.48\linewidth]{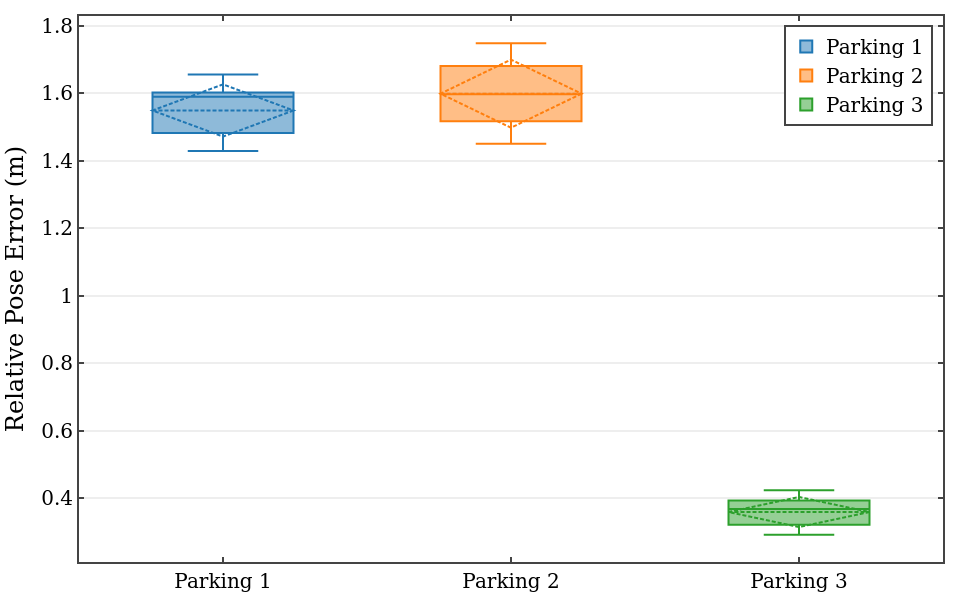}
 }
 \captionsetup[subfigure]{width=.40\linewidth}
 \subfloat[RPE comparison between bundle adjustment using PGO and Loop-box.] {\label{fig-ape-loopbox-pgo}
 \includegraphics[width=.48\linewidth]{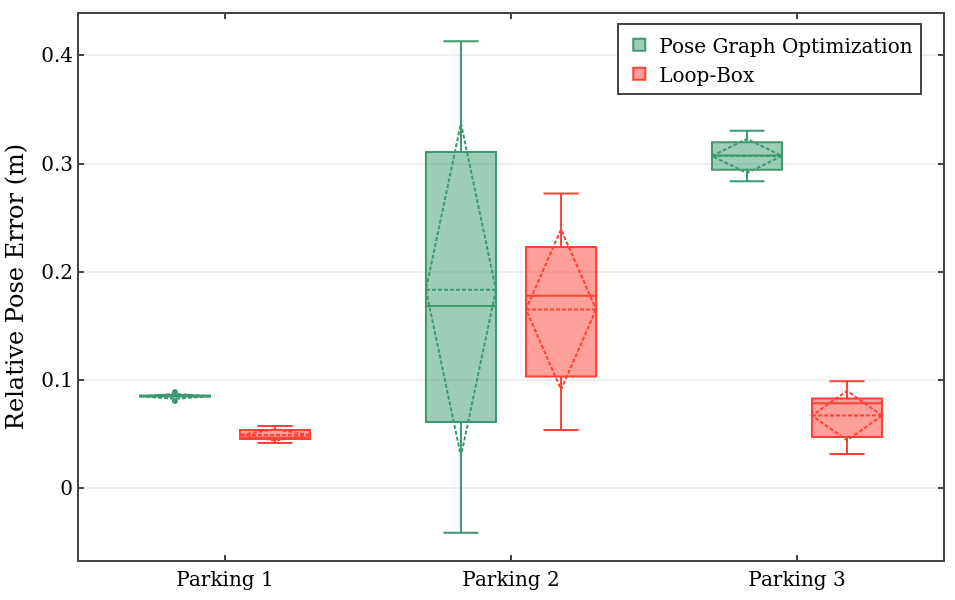}
 }
 \caption{Relative pose error (RPE) (a) PCR-Pro applied directly to the first match corresponding to each dataset. The RPE results show that the error varies due to the number of matched keypoints of each matching. (b) Evaluation of the Loop-box performance compared to state-of-the-art PGO \cite{kummerle2011g} for each dataset.}
 \label{fig-ape} 
\end{figure*}

In Table \ref{table_1}, we compare the relative pose root mean square error (RMSE) of Loop-box with that of PCR-Pro \cite{Bhutta2018} and pose graph optimization (PGO) \cite{kummerle2011g}. Please note that both the Loop-box and PGO methods use a scale difference and relative transformation firstly computed by PCR-Pro. This is why their results are better as compared to PCR-Pro's.
Furthermore, we separately analyze the relative pose error (RPE) of PCR-Pro on the three parking datasets. For \textit{Parking 1} and \textit{Parking 2}, the RPE is much higher than for the \textit{Parking 3} dataset, as shown in Fig. \ref{fig-ape-pcr-pro}.
If we look at the performance of bundle adjustment using PGO, the RPE is small. The interesting factor to note is that Loop-box outperforms PGO further. We evaluate the Loop-box performance compared to PGO's, as shown in Fig. \ref{fig-ape-loopbox-pgo}. Moreover, the estimation time of Loop-box is much lower than that of PGO, which requires additional non-linear optimization for the bundle adjustment. The computation time and RMSE are compared in Table \ref{table_1}.
To show the robustness of Loop-box according to the motivation mentioned in Section \ref{sec.introduction}, we discuss the results in two sections: uni- and bi-directional results and multi-agent 3-D mapping results.
\begin{figure*}[!t] 
 \centering
 \captionsetup[subfigure]{width=.32\linewidth}
 \subfloat[\textit{Parking 2} dataset: Map merging using PGO.]{\label{fig-loop-box-pgo-parking2-a}
 \includegraphics[width=.32\linewidth]{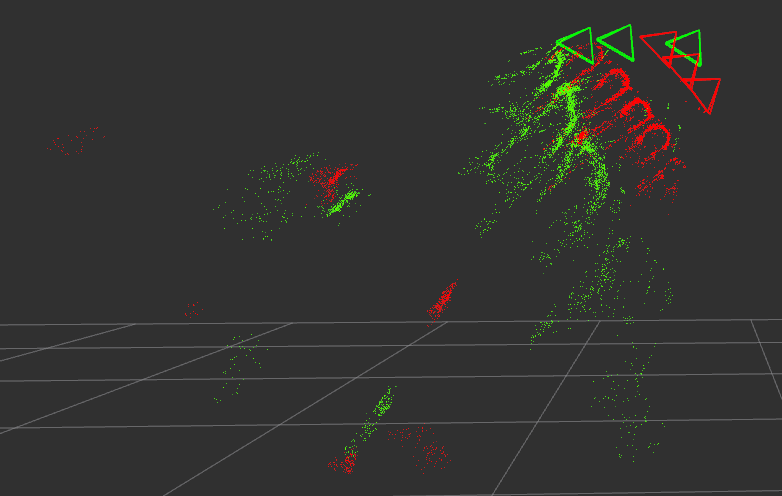}
 }
 \subfloat[\textit{Parking 2} dataset: Map merging using Loop-box.]{\label{fig-loop-box-pgo-parking2-b}
 \includegraphics[width=.32\linewidth]{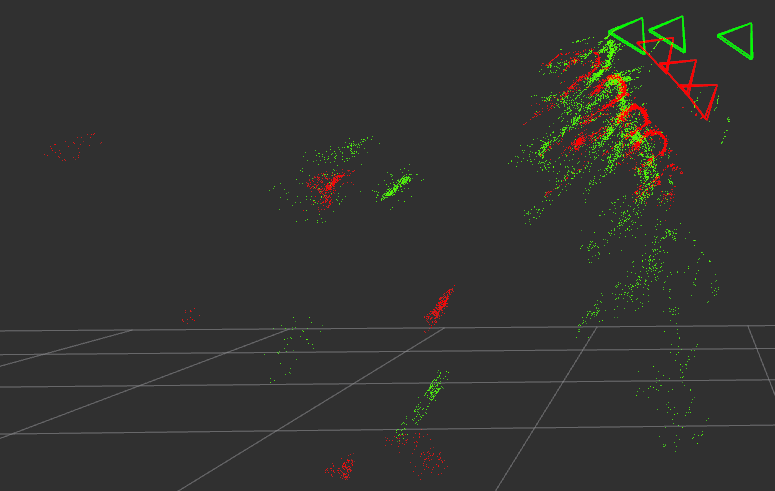}
 }
 \subfloat[\textit{Parking 2} dataset: Illustration of all poses, which are transformed by both Loop-box and PGO.]{\label{fig-loop-box-pgo-parking2-c}
 \includegraphics[width=.32\linewidth]{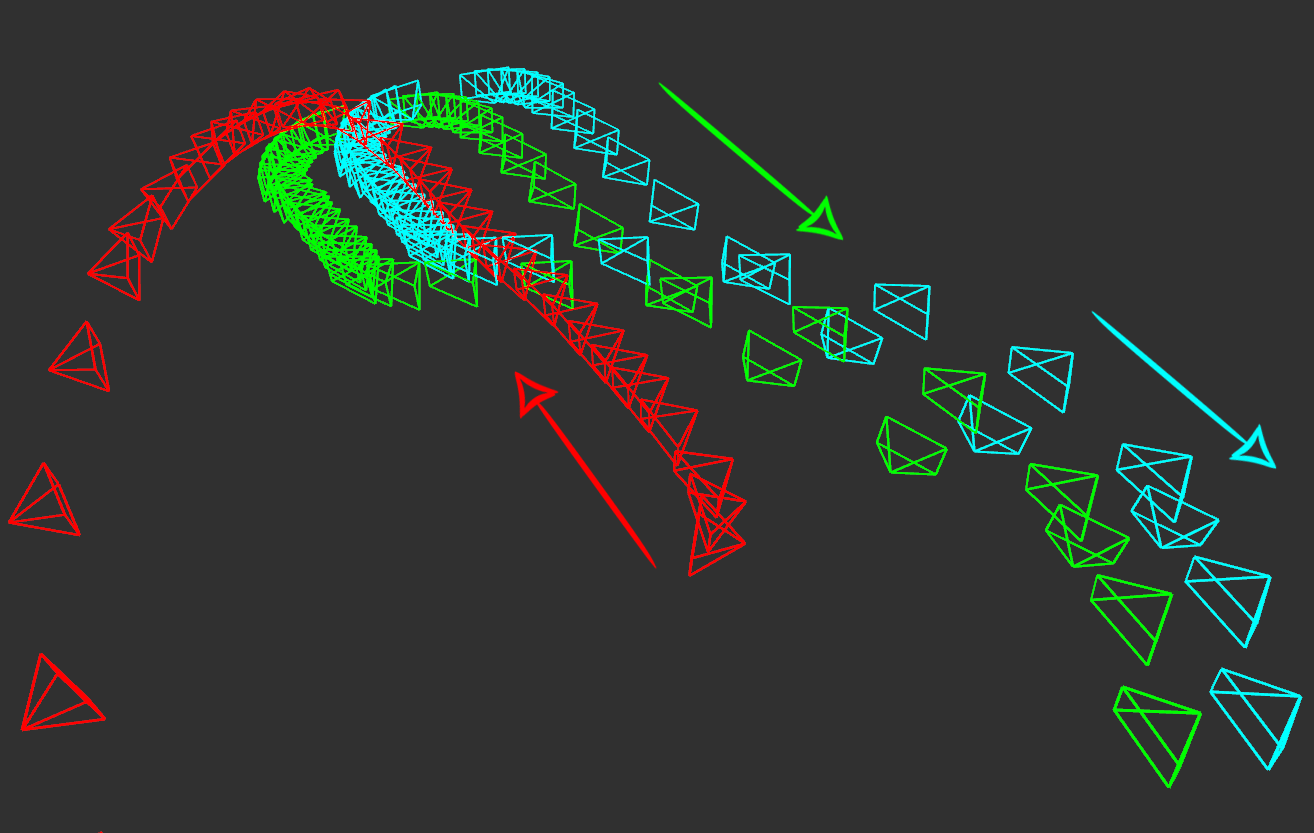}
 }\hfil
 \subfloat[\textit{Parking 3} dataset: Map merging using PGO.]{\label{ffig-loop-box-pgo-parking2-d}
 \includegraphics[width=.32\linewidth]{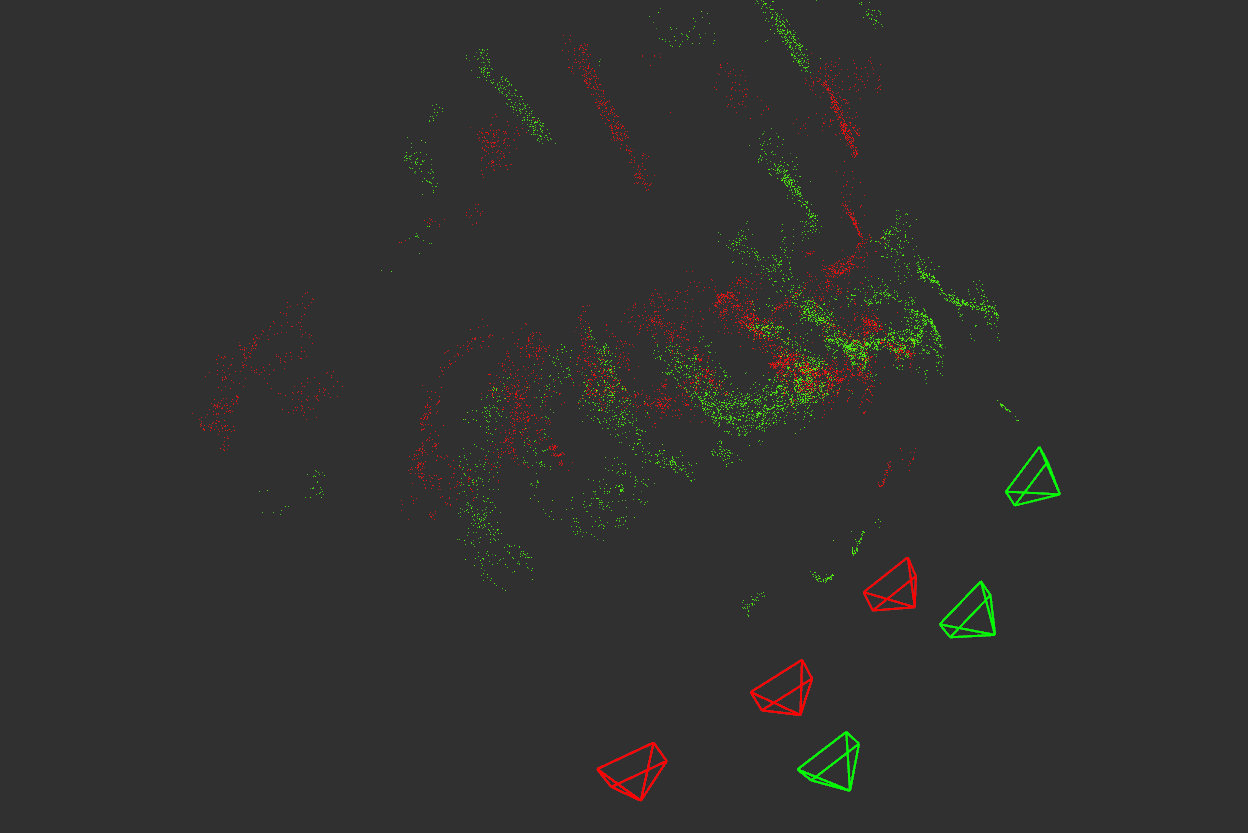}
 }
 \subfloat[\textit{Parking 3} dataset: Map merging using Loop-box.]{\label{ffig-loop-box-pgo-parking2-e}
 \includegraphics[width=.32\linewidth]{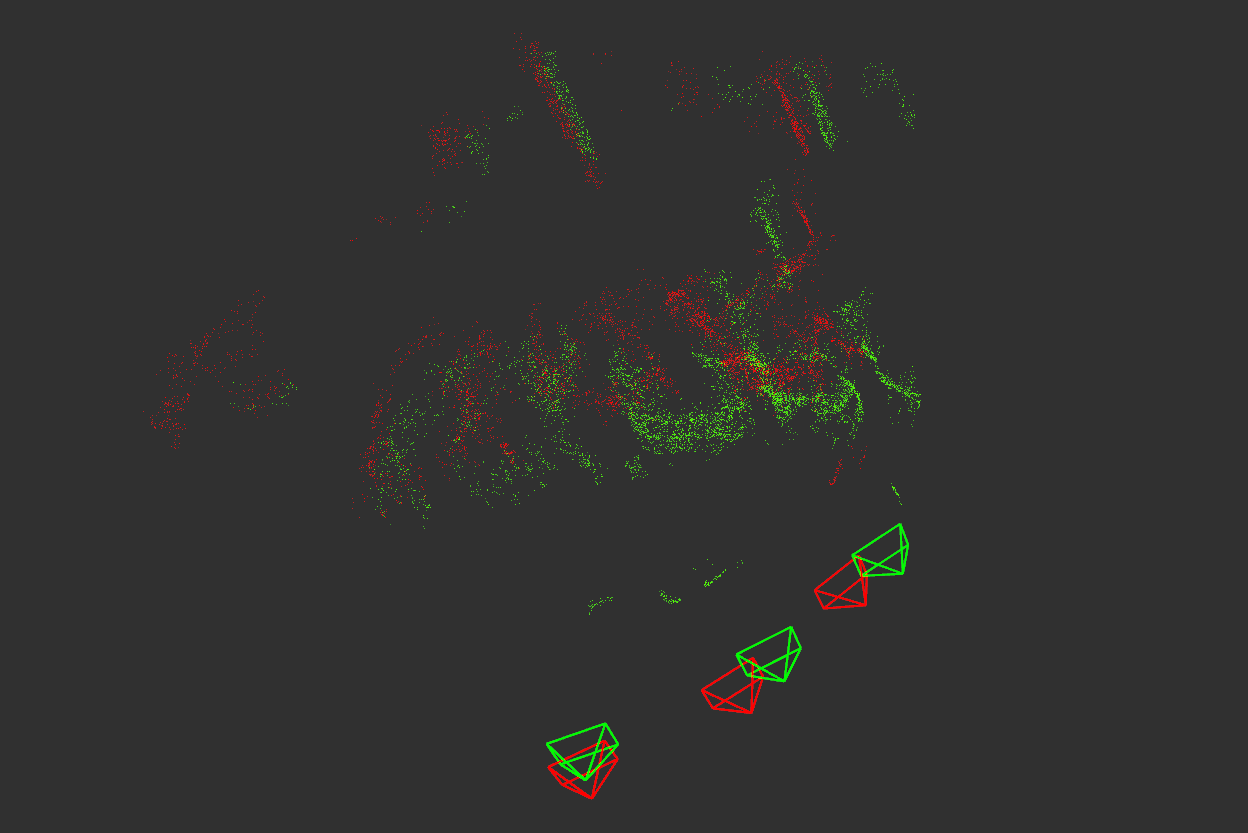}
 }
 \subfloat[\textit{Parking 3} dataset:  Illustration of all poses, which are transformed by both Loop-box and PGO.]{\label{fig-loop-box-pgo-parking2-f}
 \includegraphics[width=.32\linewidth]{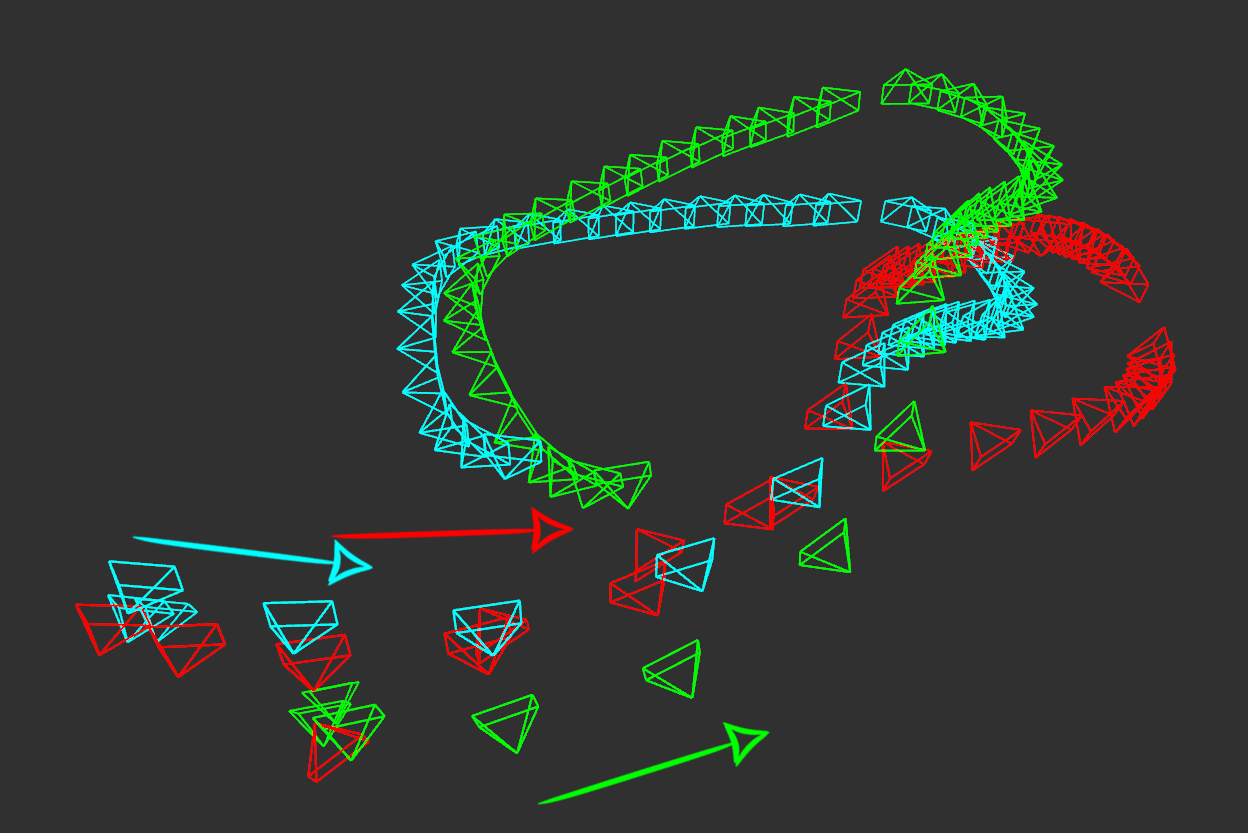}
 }
 \caption{(a), (b), and (c) correspond to the \textit{Parking 2} dataset and (d), (e), and (f) correspond to the \textit{Parking 3} dataset. Travelling directions of the agents are shown with arrows in (c) and (f). In (a), (b), (d), and (e), red represents the source poses and point clouds and green represents the target agent poses and point clouds. In (c) and (f), the source is in red, while the green camera poses are those estimated by the PGO method, and cyan represents the camera poses transformed by Loop-box.}
 \label{fig-loop-box-pgo-parking2} 
\end{figure*}
\begin{figure*}[!t] 
 \captionsetup[subfigure]{width=.48\linewidth}
 \subfloat[\textit{Parking 2} dataset: SLAM results after bundle adjustment by PGO.]{\label{fig_parking3a}
 \includegraphics[width=.49\linewidth]{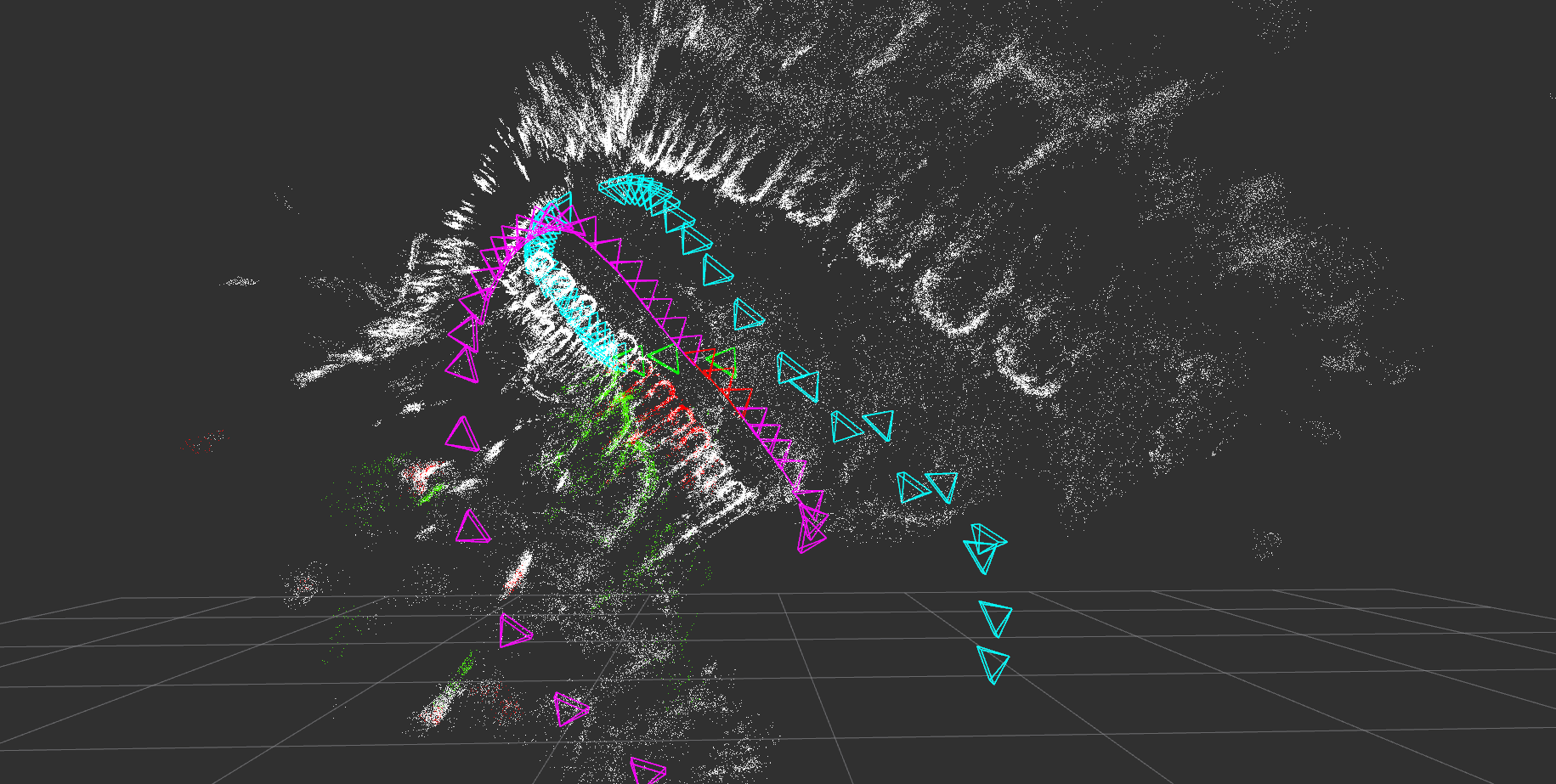}
 }
 \subfloat[\textit{Parking 2} dataset: SLAM results using Loop-box.]{\label{fig_parking3b}
 \includegraphics[width=.49\linewidth]{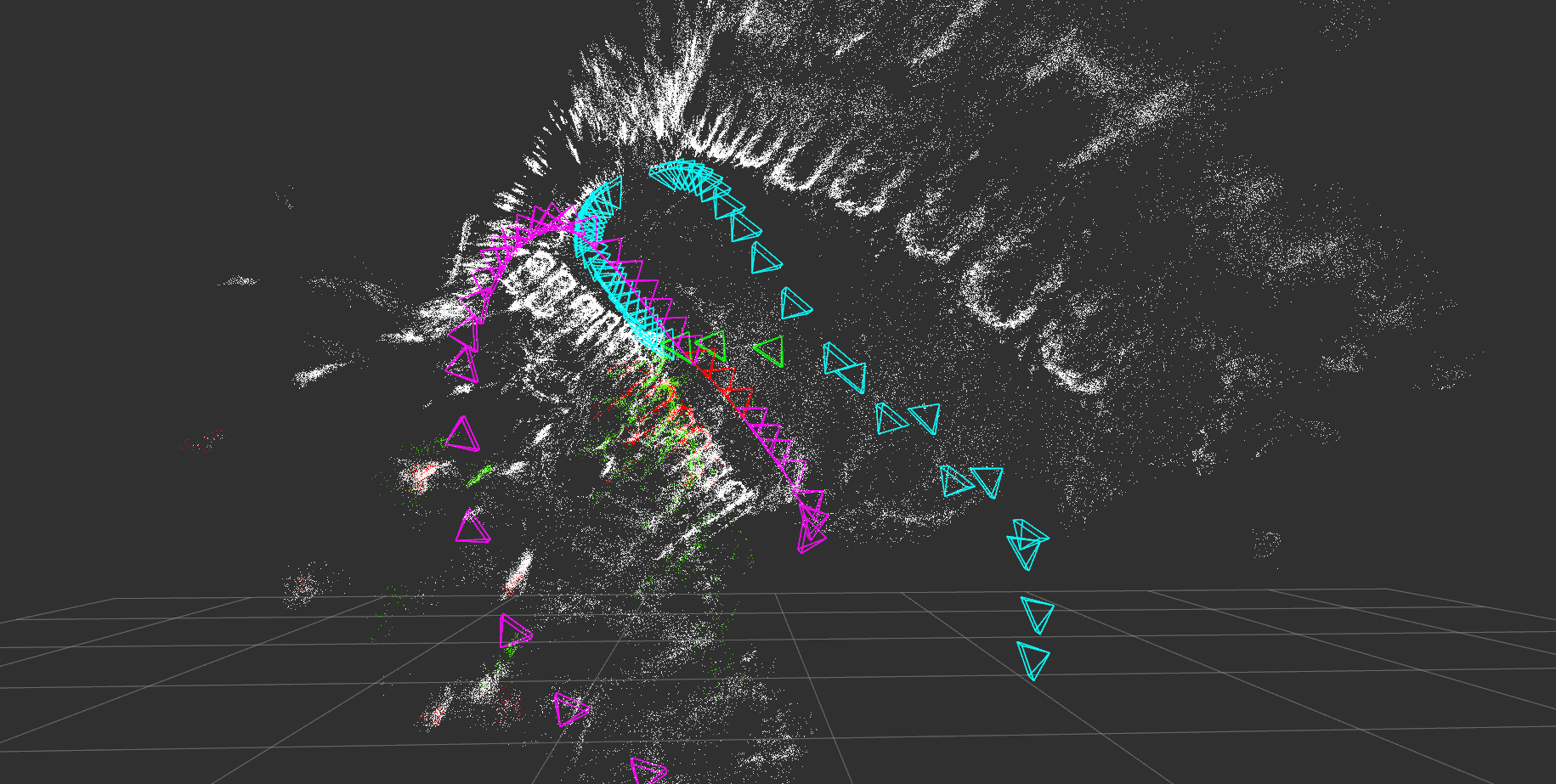}
 }
 \hfil
 \captionsetup[subfigure]{width=.48\linewidth}
 \subfloat[\textit{Parking 3} dataset: SLAM results after bundle adjustment by PGO.]{\label{fig_parking3c}
 \includegraphics[width=.49\linewidth]{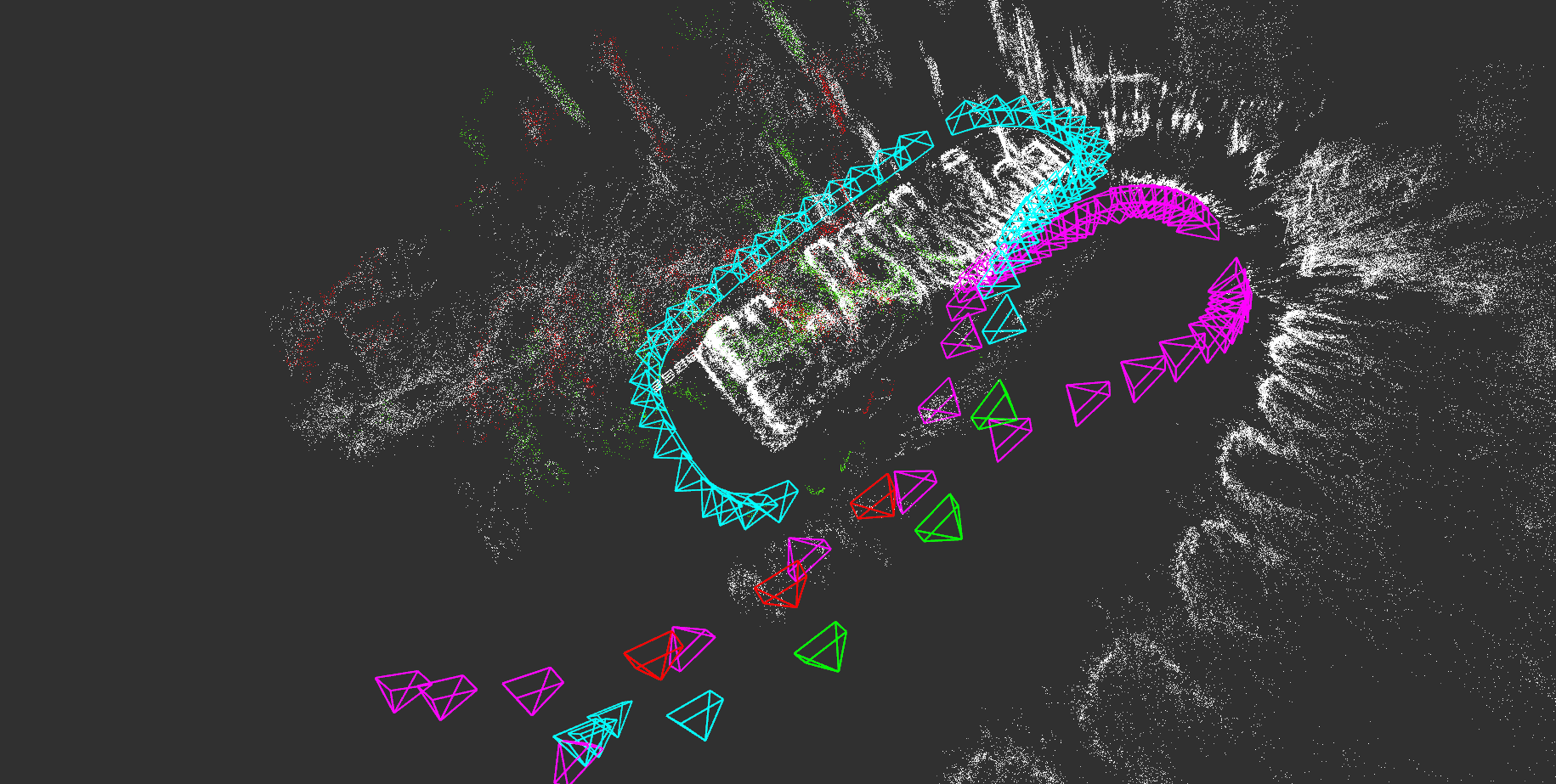}
 }
 \subfloat[\textit{Parking 3} dataset: SLAM results using Loop-box.]{\label{fig_parkingda}
 \includegraphics[width=.49\linewidth]{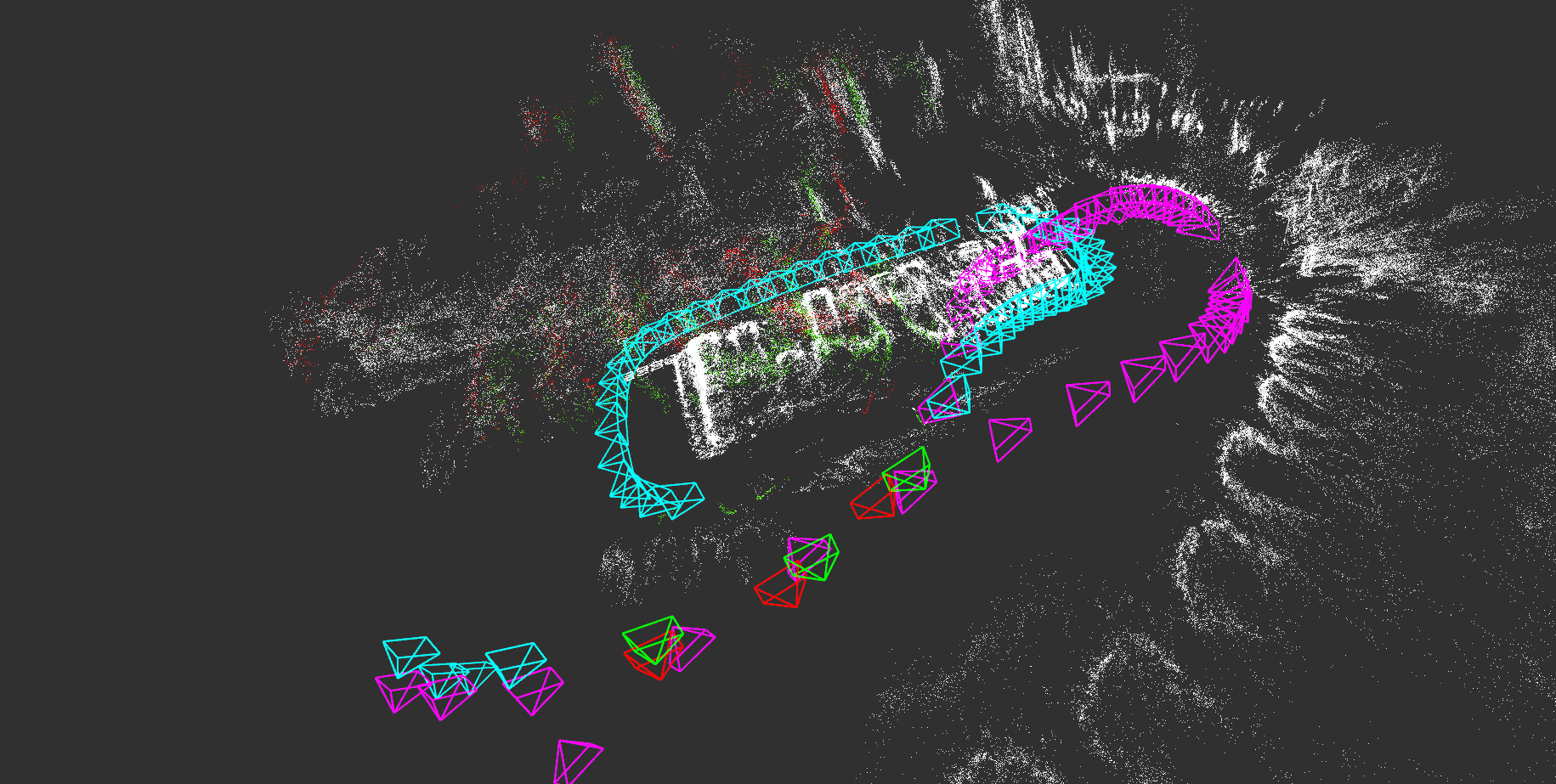}
 }\hfil
 \caption{Multi-agent SLAM including full map merging comparison of Loop-box with PGO. Poses of the source agent and target are shown in magenta and cyan, respectively. The loop closure matches of the source and target agent are shown in red and green, respectively.}
 \label{fig_parking3} 
\end{figure*}
\subsection{Uni- and Bi-Directional Results}
For the uni- and bi-directional results, the \textit{Parking 2} and \textit{Parking 3} datasets were recorded using the UGV shown in Fig. \ref{fig3b}. These datasets consist of three bag files. The first two bags have the same path direction for loop closure, and the third bag has the opposite direction. We compare the map merging results of PGO and Loop-box in Fig. \ref{fig-loop-box-pgo-parking2} since both perform better than PCR-Pro. The results show that the map merging by Loop-box converges to the global optimum in contrast to the bundle adjustment using PGO. The difference is viewed best in the matched poses location. All camera poses along with 3-D maps transformed by bundle adjustment using the PGO and Loop-box methods are shown in Fig. \ref{fig-loop-box-pgo-parking2}.
\subsection{Multi-Agent 3-D Mapping}
\label{sec.multi}
\begin{figure*}[!t] 
    \captionsetup[subfigure]{width=.48\linewidth}
    \subfloat[\textit{Parking 2} and \textit{Parking 3} datasets: PGO-based multi-agent results.]{\label{fig_multi_agentsa}
        \includegraphics[width=.49\linewidth]{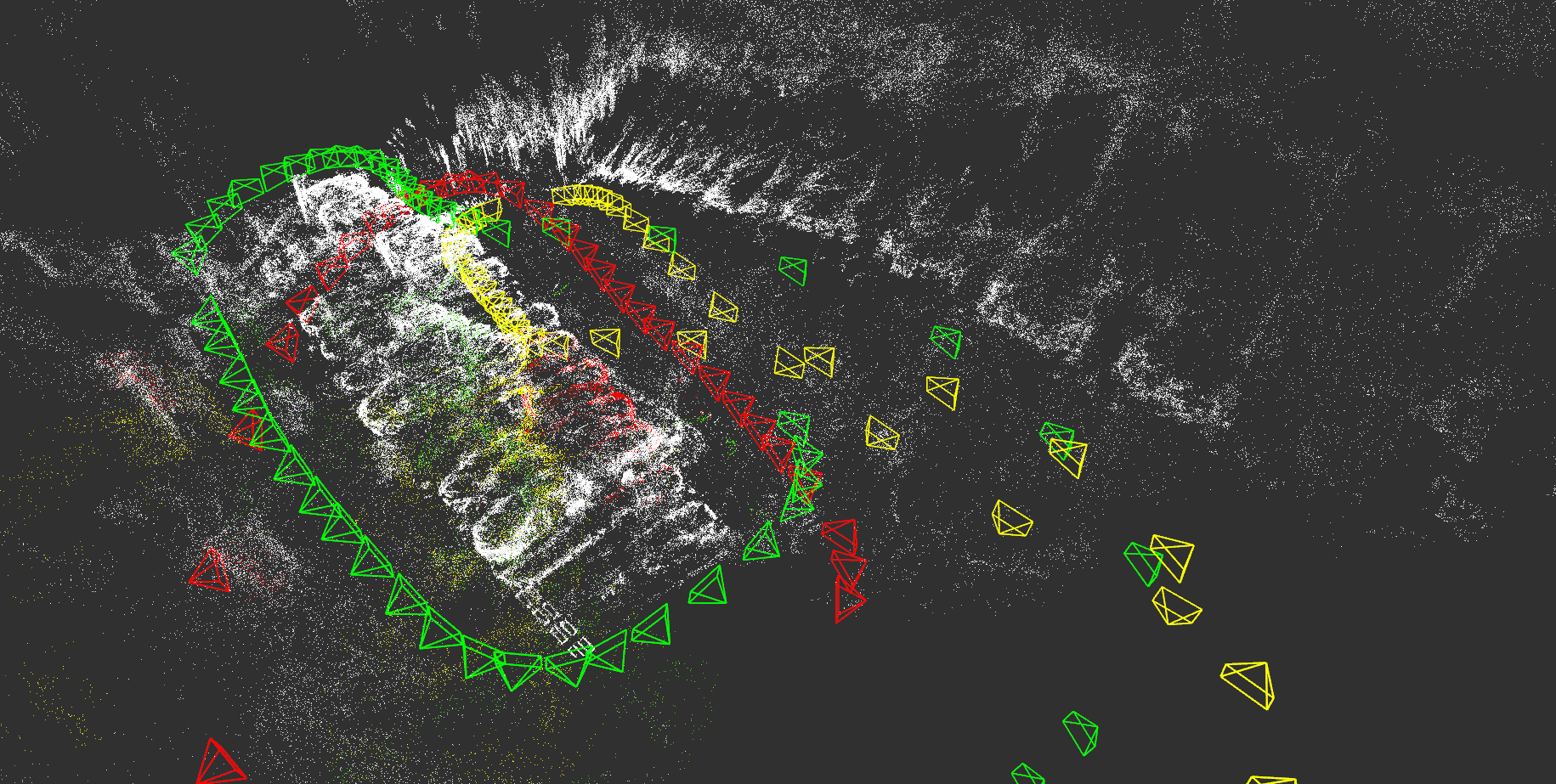}
    }
    \subfloat[\textit{Parking 2} and \textit{Parking 3} datasets: Loop-box-based multi-agent results.]{\label{fig_multi_agentsb}
        \includegraphics[width=.49\linewidth]{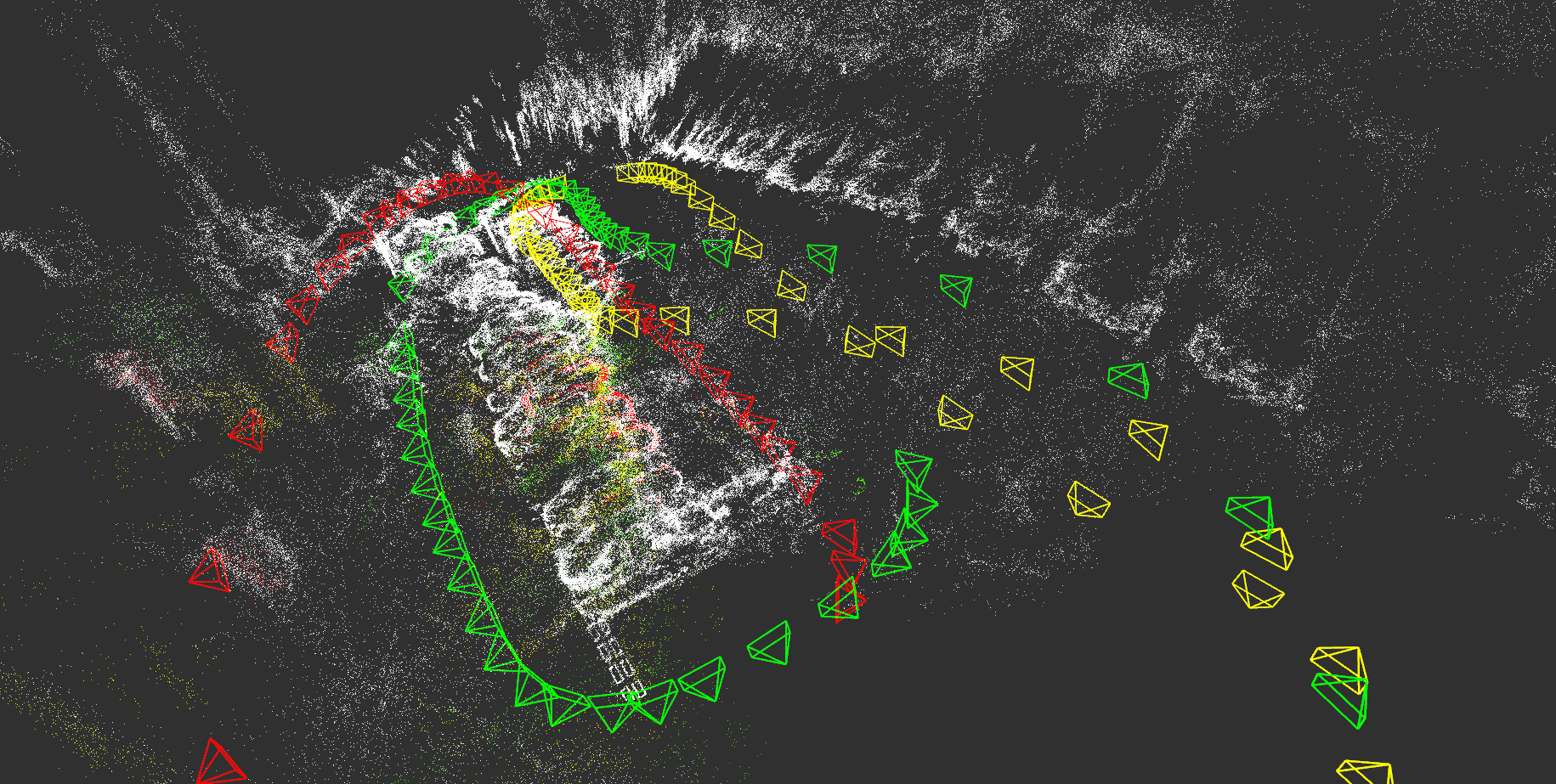}
    }
    \caption{
Multi-agent 3-D map fusion. The target agent of the \textit{Parking 2} dataset is the source agent of the \textit{Parking 3} dataset. The odometry of this connecting agent is shown in yellow. The source agent of the \textit{Parking 2} dataset and target agent of the \textit{Parking 3} dataset are shown in red and green, respectively.}
    \label{fig_multi_agents} 
\end{figure*}

To have multi-agent 3-D mapping results, we combined the results of the \textit{Parking 2} and \textit{Parking 3} datasets. In the \textit{Parking 3} dataset, both agents are moving in the same direction, whereas in the \textit{Parking 2} dataset, the agents are moving in opposite directions. If the system can detect merely one loop closure, then the 3-D maps can easily be merged using Loop-box. All three agents originally had different scales and camera odometries. After estimating the optimal scale, we processed the three adjacent matches using Loop-box and PGO, as shown in Fig. \ref{fig_parking3}. Since the target agent of the \textit{Parking 2} dataset is the source agent of the \textit{Parking 3} dataset, we name the agent a connecting agent. 
Firstly, we scale the source agent of the \textit{Parking 2} dataset according to the connecting agent. This transformation was further scaled according to the target agent of the \textit{Parking 3} dataset. Next, the target agent of the \textit{Parking 3} dataset was transformed according to the merged \textit{Parking 2} dataset. We tested Loop-box compared to the PGO method to merge the 3-D map with the connecting agent. The full map merging results are shown in Fig. \ref{fig_multi_agents}.
Material related to this work is available at \url{https://usmanmaqbool.github.io/loop-box}.

\section{Discussion}
\label{sec.discussions}
A challenging application of the multi-agent SLAM system is a crowd-based mapping system where different people use their camera phones to capture the 3-D environment and then collaborate to yield a large-scale 3-D mapping. After recognizing the first loop closure, our Loop-box framework is successfully able to find the relative pose transformation regardless of the scale difference between them. For the PCR-Pro results, we estimated the relative transformation individually. In the pose graph optimization, we selected the top three methods, which are discussed in detail in Appendix \ref{sec.posegraph}. In the end, we used our proposed method, which needs only one match out of three based on the keypoints. For all the datasets, 3-D map merging accuracy describes the efficiency of the Loop-box method where the system converges to the global minimum, as presented in Table \ref{table_1}.
We discuss the limitation of multi-agent SLAM systems and specifications for improving Loop-box as follows. 
\subsection{Place Recognition}
With the short-interval loop closure, a number of scenarios can happen. If both agents are moving in the same direction, the system does not need any modification. But if both agents are coming from opposite directions, as in the \textit{Parking 2} dataset, the camera system should capture the side view of the agent path so both agents can accurately detect the loop closure. After the loop closure, the adjacent keyframe matching will not be in a direct but in a crossed manner, as shown in red in Fig. \ref{fig3a}. For the distributed multi-agent SLAM system, if two agents detect that they are in range of each other, they can quickly establish relative pose connection using Loop-box.\\
In this work, we use FABMAP\cite{cummins2011appearance} and create the visual vocabulary for loop closure detection. Preparing a visual vocabulary is time-consuming and even not recommended for large-scale mapping. Some methods \cite{galvez2012bags} also offer a pre-built dictionary for place recognition. This enables the system to be used without training the environment. Loop-box can be further improved by incorporating NetVLAD features\cite{arandjelovic2016netvlad}. For all, the features will be calculated and uploaded to the server instead of all the keyframes. This can give a significant improvement in the detection time of loop closures. The NetVLAD features handling is also simpler than vocabulary handling. Sometimes, biologically inspired visual SLAM systems \cite{milford2010persistent,ni2016bioinspired} also help in persistent navigation and mapping in a dynamic environment.
\subsection{Drift}
In our results, we can see excellent transformations at the loop closure area, but a small drift is clearly visible afterward. Since LSD-SLAM is based on a monocular camera only, drift can affect local mapping. The Loop-box method allows monocular camera-based SLAM systems to connect and find the exact transformation. These systems can have drift problems, which can be solved using additional sensors. For example, OKVIS can track a local map built from several recently captured keyframes, which significantly minimizes the local drifts \cite{Leutenegger2015}. It can also be improved by a tightly coupled configuration for sensor fusion of the camera with an IMU. This helps in avoiding the local drift in the visual odometry of the agent.
\section{Conclusions}
\label{sec.conclusions}
This paper has presented a framework that can estimate real-time 3-D relative transformation of different agents for large-scale SLAM and 3-D map-merging applications. The results have shown admirable performance regarding map merging and accurate estimation of the relative pose transformation after only a single loop closure. After the first matches, our system has the ability to localize each agent with respect to the global reference frame while keeping track of and continuously adding keyframes and point cloud data to the global map. This work can be extended to \textit{n} agents of any kind, for instance, unmanned air and ground vehicles. Features of this research are its computationally robust framework, and its possible applicability to swarm robotics in map exchange, and task allocation applications. 
\appendices
\section{Bundle Adjustment Study for Multi-Agent SLAM} 
\label{sec.posegraph}
Several studies suggest that bundle adjustment using PGO performs excellently in SLAM systems. To compare Loop-box with bundle adjustment, we additionally estimated the covariance among edges. PCR-Pro \cite{Bhutta2018} along with libpointmatcher \cite{Pomerleau2011} were used to achieve high-level optimal transformation. Using the estimated covariance based on the sets of correspondence and relative transformation, we can easily determine the information matrix, enabling the system for PGO.
We considered three different cases, and applied the PGO to the edge transformations, which are shown by the grey dotted lines in Fig. \ref{fig_BA}.
\subsubsection{Straight Configuration}
For the straight configuration, we examined directly matched camera poses, as shown in Fig. \ref{ba-1}. $ S1 $, $ S2 $, and $ S3 $ are source poses, while $ T1 $, $ T2 $, and $ T3 $ are the corresponding target poses. We applied the transformation $ ^{s1}\mathbf{T}_{t1}^{*} $, $ ^{s2}\mathbf{T}_{t2}^{*} $, and $ ^{s3}\mathbf{T}_{t3}^{*} $ estimated in Equation \ref{eqn10} to the $ T1 $, $ T2 $, and $ T3 $ poses, respectively. Moreover, we scaled the $ S1 $, $ S2 $, and $ S3 $ poses by $ ^{\sigma}\mathbf{T}_{s1}^{*}$, $ ^{\sigma}\mathbf{T}_{s2}^{*} $, and $ ^{\sigma}\mathbf{T}_{s3}^{*} $. Then we applied the bundle adjustment for overall transformation optimization.
\subsubsection{Fully Connected Configuration}
\begin{figure}[!t] 
 \captionsetup[subfigure]{width=.48\linewidth}
 \subfloat[Direct match configuration]{\label{ba-1}
 \includegraphics[width=.49\linewidth]{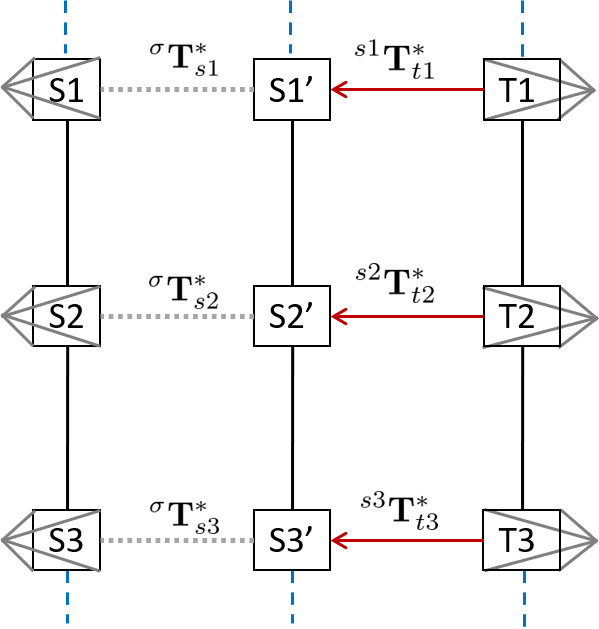}
 }
 \subfloat[Fully connected configuration]{\label{ba-2}
 \includegraphics[width=.49\linewidth]{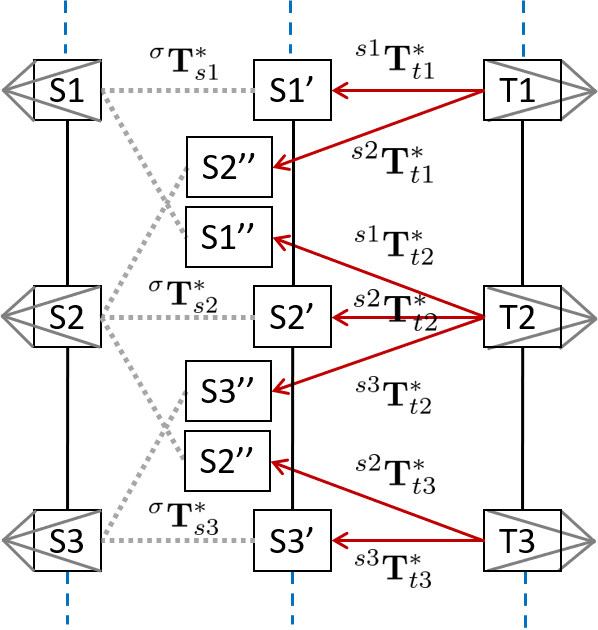}
 }
 \caption{Different pose configurations for PGO.}
 \label{fig_BA}
\end{figure}

After the loop closure, cross matched pairs $\{T1,S2''\}$, $\{T2,S1''\}$, $\{T2,S1''\}$, $\{T2,S3''\}$, and $\{T3,S2''\}$ of three adjacent poses were also selected along with the directly matched pairs for the full PGO, as shown in Fig. \ref{ba-2}.
\subsubsection{Top Matches Configuration}
This configuration is similar to the straight configuration as explained above. The difference is that we use a single transformation $ ^{s(j)}\mathbf{T}_{t(i)}^{*} $, where $i$ to $j$ correspond to those poses that have large matched keypoints $ \gamma $. For two agents, there is one optimal transformation for all the poses.

In all the above configurations, the top matches configuration gives much better results than the direct and fully connected setup. After the PGO using this best pose solution, we use the middle poses of $ S2 $ and $ T2$. Another good factor in using this for bundle adjustment is that it not only estimates better relative transformation, but also takes less time in computation. The top matches configuration results have been compared with the Loop-box method in Section \ref{sec.experimental_results}.

\bibliographystyle{IEEEtran}
\bibliography{paper} 

\end{document}